\documentclass[10pt,twocolumn,letterpaper]{article}

\usepackage{cvpr}              

\usepackage{comment}
\usepackage{multirow}
\usepackage{algorithm,algpseudocode}
\usepackage{url}
\usepackage[normalem]{ulem}
\usepackage{bbding}
\usepackage{xspace}
\usepackage{tabu}%
\usepackage{longtable}
\usepackage[utf8]{inputenc}
\usepackage[T1]{fontenc}
\usepackage{siunitx}

\usepackage{amsfonts}
\usepackage{amsmath}
\usepackage{pifont}
\newcommand{\xmark}{\ding{55}}%
\usepackage{amssymb}
\usepackage{amsthm}
\usepackage{setspace}
\usepackage{array}
\usepackage{bbm}
\usepackage{tabularx}
\usepackage{calc}
\usepackage{makecell}
\usepackage[accsupp]{axessibility} 
\usepackage{subcaption,graphicx}

\usepackage{booktabs}  

\usepackage[table,x11names]{xcolor}
\definecolor{mygray}{gray}{0.5}

\usepackage{caption} 
\usepackage[pagebackref,colorlinks]{hyperref}
\usepackage[capitalize]{cleveref}
\crefname{section}{Sec.}{Secs.}
\Crefname{section}{Section}{Sections}
\Crefname{table}{Table}{Tables}
\crefname{table}{Tab.}{Tabs.}

\newcommand{\ours}{Leopart}
\newcommand\blfootnote[1]{%
  \begingroup
  \renewcommand\thefootnote{}\footnote{#1}%
  \addtocounter{footnote}{-1}%
  \endgroup
}

\newcommand{\midsepchange}{\aboverulesep = 0.05mm \belowrulesep = 0.2mm}
\midsepchange

\def\cvprPaperID{5807} 
\def\confName{CVPR}
\def\confYear{2022}

\makeatletter
\renewcommand{\paragraph}{%
  \@startsection{paragraph}{4}%
  {\z@}{0.75ex \@plus 1ex \@minus .2ex}{-1em}%
  {\normalfont\normalsize\bfseries}%
}
\makeatother

\begin{document}

\newpage
\title{Self-Supervised Learning of Object Parts for Semantic Segmentation}
\author{Adrian Ziegler\\
Technical University of Munich  \\
{\tt\small adrian.ziegler@tum.de}
\and
Yuki M. Asano\\
QUVA Lab \\
University of Amsterdam\\
{\tt\small y.m.asano@uva.nl}
}
\maketitle

\begin{abstract}
Progress in self-supervised learning has brought strong image representation learning methods. 
Yet so far, it has mostly focused on image-level learning.
In turn, tasks such as unsupervised image segmentation have not benefited from this trend as they require spatially-diverse representations.
However, learning dense representations is challenging, as in the unsupervised context it is not clear how to guide the model to learn representations that correspond to various potential object categories.
In this paper, we argue that self-supervised learning of object parts is a solution to this issue.
Object parts are generalizable: they are a-priori independent of an object definition, but can be grouped to form objects a-posteriori.
To this end, we leverage the recently proposed Vision Transformer's capability of attending to objects and combine it with a spatially dense clustering task for fine-tuning the spatial tokens.
Our method surpasses the state-of-the-art on three semantic segmentation benchmarks by 3\%-17\%, showing that our representations are versatile under various object definitions.
Finally, we extend this to fully unsupervised segmentation -- which refrains completely from using label information even at test-time -- and demonstrate that a simple method for automatically merging discovered object parts based on community detection yields substantial gains. \blfootnote{Code: \url{https://github.com/MkuuWaUjinga/leopart}}

\end{abstract}

\begin{figure}[!t]
    \centering
    \includegraphics[width=\columnwidth]{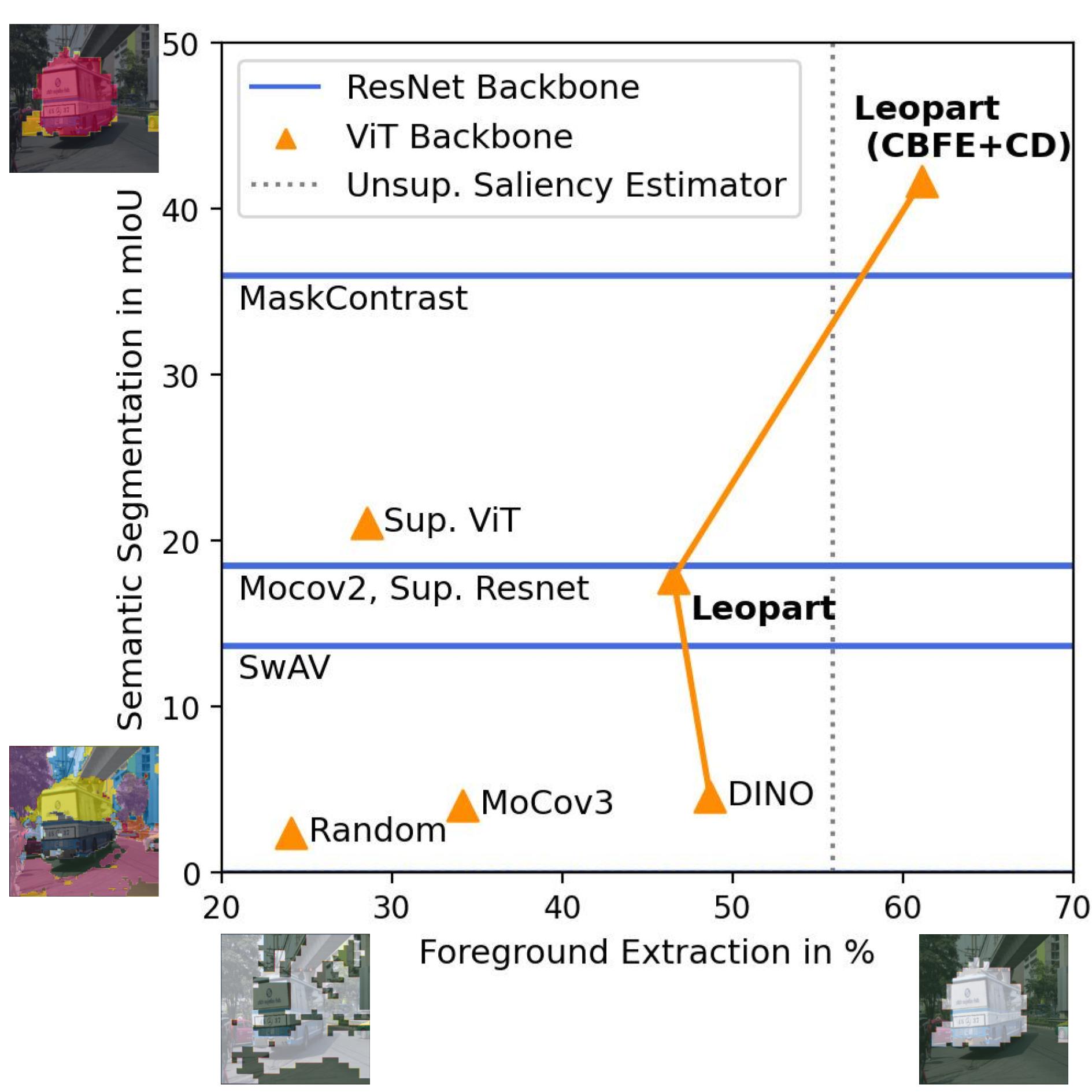}
    \caption{\textbf{ViTs and Resnets compared under foreground extraction and semantic segmentation.} We use Jaccard distance as a measure for foreground extraction. Starting from a DINO initialization, our method, \ours, closes the performance gap between self-supervised ViTs and their supervised counterparts as well as Resnets. \ours~(CBFE+CD) further improves a ViT's object extraction capabilities and sets new state-of-the-art for fully unsupervised semantic segmentation. 
    }
    \label{fig:splash}
\end{figure}
\section{Introduction}\label{s:intro}
Defining what makes an object an object is hard.
In philosophy, Peirce defines an object as anything we can think and talk about~\cite{peirceobject}.
In computer vision, object definitions for semantic segmentation are more pragmatic and feature various notions of objectness as well as different levels of granularity.
For instance, the COCO-Stuff benchmark distinguishes between stuff (objects without a clear shape) and things (objects with a ``well-defined'' shape)~\cite{caesar2018coco, lin2014microsoft} and features coarse and fine object categories.
Others, like Cityscapes~\cite{Cordts2016Cityscapes}, choose a segmentation that is most informative for a specific application like autonomous driving and therefore also include sky and road as object classes.

This variedness of object definitions is challenging for self-supervised or unsupervised semantic segmentation as human annotations that carry the object definitions are, at most, used at test time.
However, the ability to learn self-supervised dense representations is desirable as this would allow scaling beyond object-centric images and allow effective learning on billions more generic images.
Furthermore, unsupervised segmentation can be highly useful as a starting point for more efficient data labeling, as segmentation annotations are even more expensive than image labelling~\cite{lin2014microsoft}.
To tackle the lack of a principled object definition during training, many methods resort to defining object priors such as saliency and contour detectors to induce a notion of objectness into their pretext tasks~\cite{cho2015unsupervised, hwang2019segsort, zhang2020self, vangansbeke2020unsupervised, henaff2021efficient}, effectively rendering such methods semi-supervised and potentially not generalizeable.
In this paper, we instead stay in the fully unsupervised domain and explore a novel, yet simple alternative for training densely.
We \textbf{le}arn \textbf{o}bject \textbf{part}s (Leopart) through a dense image patch clustering pretext task.
Object part learning promises a principled formulation for self-supervised dense representation learning as object parts can be composed to form objects as defined in each benchmark, after generic pretraining.

In this paper, we explore the use of a Vision Transformer (ViT) with our new loss and excavate its unique aptness for self-supervised segmentation.   
While vision transformers have shown great potential unifying architectures and scaling well with data into billions, they have mostly been shown to work for image-level tasks~\cite{dosovitskiy2021image, chen2021empirical, caron2021emerging} or dense tasks~\cite{wang2021pyramid, wu2021cvt, liu2021swin, chu2021twins} but in a supervised manner.
Our work aims to close this gap by self-supervisedly learning dense ViT models.
We combine the recently discovered property of self-supervised ViTs to localise objects~\cite{caron2021emerging} with our dense loss to train spatial tokens for unsupervised segmentation.

We validate our method from two different angles: First, we conduct a transferability study and show that our representations perform well on downstream semantic segmentation tasks.
Second, we tackle the more challenging \textit{fully} unsupervised setup proposed in \cite{vangansbeke2020unsupervised} based on directly clustering the pixel or patch embeddings.
For that, two model characteristics are important: unsupervised foreground extraction and a semantically-structured embedding space, see Figure \ref{fig:splash}.
To our surprise, even though self-supervised ViTs excel at extracting objects, they do not learn a spatial token embedding space that is discriminative for different object categories. 
On the other hand, ViTs trained under supervision achieve better semantic segmentation performance, but the attention heads perform poorly at localizing objects.
In contrast, our method outperforms self-supervised ViTs \textit{and} ResNets in fully unsupervised semantic segmentation as well as in learning transferable dense representations.
\par
Thus, our contributions are as follows:
\begin{itemize}
\itemsep-0.2em 
    \item We propose a dense clustering pretext task to learn semantically-rich spatial tokens, closing the gap between supervised ViTs and self-supervised ViTs.
    \item We show that our pretext task yields transferable representations that surpass the state-of-the-art on Pascal VOC, COCO-Thing and COCO-Stuff semantic segmentation  \textit{at the same time} by 17\%-3\%.
    \item We develop a novel cluster-based foreground extraction and overclustering technique based on community detection to tackle fully unsupervised semantic segmentation and surpass the state-of-the-art by >6\%.
\end{itemize}

\section{Related Work}
Our work takes inspiration from standard image-level self-supervised pretraining while extending this to the domain of dense representation learning using Vision Transformers.

\paragraph{Image-level self-supervised learning.}
Self-supervised learning aims to learn powerful representations by replacing human annotation with proxy tasks derived from data alone.
Current methods can be roughly categorized into instance-level and group-level objectives.
Instance-level objectives include predicting augmentations applied to an image \cite{pathak2016context, Doersch_2015_ICCV, noroozi2016unsupervised, zhang2016colorful, jenni2018selfsupervised, zhang2019aet, gidaris2018unsupervised}
or learning to discriminate between images \cite{bojanowski2017unsupervised, dosovitskiy2015discriminative, Wu_2018_CVPR,he2019momentum, chen2020generative}, often done by the use of contrastive losses \cite{Hadsell2006DimensionalityRB}. 

On the other hand, group-level objectives explicitly allow learning shared concepts between images by leveraging clustering losses \cite{caron2018deep, asano2020self, van2020scan, caron2020unsupervised}.
\cite{caron2018deep} proposes k-means clustering in feature space to produce pseudo labels for training a neural network.
\cite{asano2020self} casts the problem of finding pseudo labels as an optimal transport problem unifying the clustering and representation learning objectives. 
This formulation was adapted to an online setting in SwAV~\cite{caron2020unsupervised} together with a new multi-crop augmentation strategy, a random cropping method that distinguishes between global and local crops of an image.
The IIC method~\cite{ji2018invariant}, also conducts clustering, however using a mutual information objective.
While it can also be used densely, it has been found to focus on lower-level features specific to each dataset~\cite{vangansbeke2020unsupervised}.
Another recent line of works completely refrains from group level clustering or instance-based discrimination by predicting targets from a slowly moving teacher network \cite{grill2020bootstrap, caron2021emerging}.

Our work adapts this teacher-student setup and shows its benefits beyond image-level tasks.
To this effect, we build on the clustering pretext task from~\cite{caron2020unsupervised} and reformulate it such that it can be used on an image patch level with teacher-student setups.
We also use the multi-crop augmentation strategy and provide an interpretation from the perspective of dense prediction tasks.

\paragraph{Dense self-supervised learning.}
Based on the observation that image-level learning does not imply expressive dense representations~\cite{he2018rethinking, purushwalkam2020demystifying}, 
dedicated self-supervised dense representation learning has attracted a lot of attention~\cite{ji2018invariant, hwang2019segsort, hung2019scops, Zhang_2020, vangansbeke2020unsupervised, gidaris2021obow, wang2021dense, roh2021spatially, liu2021selfemd, henaff2021efficient, li2021dense, gao2021largescale, choudhury21unsupervised, simeoni2021localizing}.
DenseCL reformulates the contrastive objective used in MoCo~\cite{he2019momentum} to work on spatial features by establishing dense correspondences accross views and is currently the state-of-the-art in transfer learning for semantic segmentation on PVOC~\cite{li2021dense}.
Other methods resort to defining an unsupervised object prior such as region proposals~\cite{cho2015unsupervised}, contour detectors~\cite{hwang2019segsort, Zhang_2020}, saliency~\cite{vangansbeke2020unsupervised} or object masks~\cite{henaff2021efficient}.
For instance, the current state-of-the-art for unsupervised semantic segmentation, MaskContrast \cite{vangansbeke2020unsupervised}, uses a pretrained saliency estimator to mine positive and negative pixels for contrastive learning.

Concurrent to our work, \cite{li2021dense} proposes an intra-image clustering step of pixel embeddings before applying a contrastive loss on the identified pixel groups to segment images. 
However, as they are reliant on combining the former with an image-level loss, it is not well-suited for more generic images with multiple objects where image-level and pixel-level semantics do not match.
In contrast, our method uses a single clustering objective that is generalized for the dense setting, but that also works on object-centric images.
Furthermore, by leveraging  ViT's natural ability to direct its attention to objects, we do not require any external saliency generator like~\cite{vangansbeke2020unsupervised}. 

There are also works that have explored unsupervised object parts segmentation~\cite{hung2019scops,choudhury21unsupervised}, with the explicit goal to determine parts given object masks.
However, our goal is different as we use part representations as an intermediary \textit{for} semantic segmentation on classic, object-level settings.

Superficially similar to our work is also another concurrent work~\cite{simeoni2021localizing}, which tackles object \textit{detection} by using the similarity between DINO's frozen last layer self-attention patch keys as a metric for merging image patches to objects.
In contrast, we use DINO's \textit{spatial tokens} and propose to fine-tune them for \textit{semantic segmentation}.  
\section{Method}
\label{sec:method}
\begin{figure}
    \centering
    \includegraphics[width=\columnwidth]{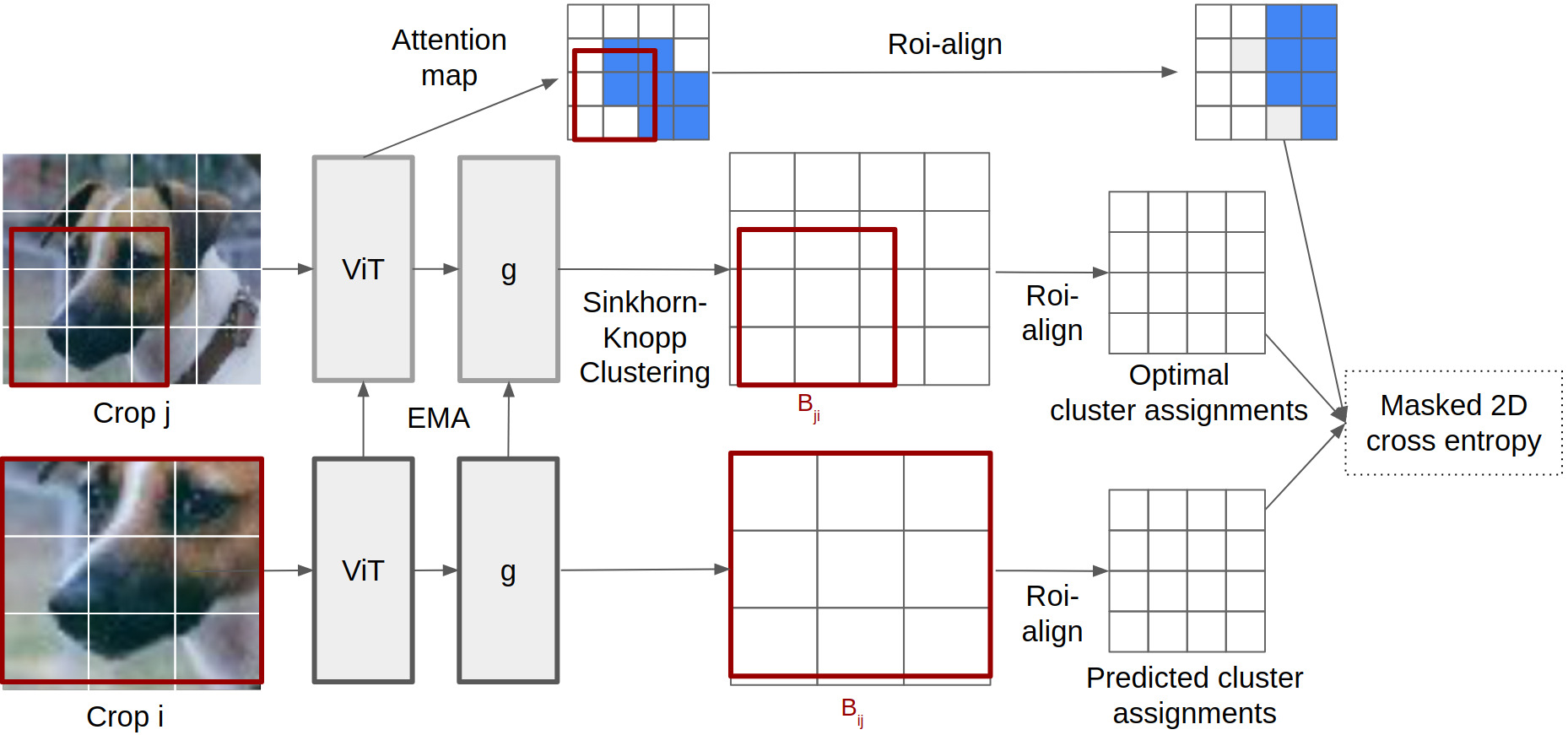}
    \caption{\textbf{Leopart training pipeline.} We start from a DINO initialization. We feed different crops to the student and teacher network to produce patch-level cluster predictions and optimal cluster assignments targets.
    This requires an alignment step of cluster targets and assignments.
    We further focus clustering on foreground patches by leveraging the ViT's attention map.}
    \label{fig:method}
\end{figure}
Our goal is to learn an embedding space that groups image patches containing the same part of an object.
This is motivated by the hypothesis that object part representations are more general than object representations, as parts can be recombined in multiple ways to different objects.
As an example, a wheel representation can be combined to a car representation but also a bus representation.
Therefore, object part representations should transfer better across datasets.
For that, we aim to design a pretext task that allows for intra-image category learning on an image-patch-level.
Thus, a clustering pretext task is a natural choice.
As shown in Figure \ref{fig:method}, we retrieve patch-level optimal cluster assignments from a teacher network and predict them from the student network. 
The choice of a clustering pretext task is further supported by empirical evidence showing that clustering pretext outperforms contrastive pretext for dense prediction tasks \cite{gao2021largescale, li2021dense}.
Instead of pretraining models from scratch which requires substantial GPU budgets, we use our loss to fine-tune pretrained neural networks. 
Further, this circumvents known cluster stability issues and clusters capturing low-level image features when applied to a patch-level as reported in~\cite{van2020scan}.

\subsection{Fine-tuning loss for spatial tokens}
\label{sec:loss-method}

\paragraph{Image Encoder.} Given an image $x\in\mathbb{R}^{3\times H \times W}$, we flatten the image into $N = \left \lfloor{\frac{H}{P}}\right \rfloor \cdot \left \lfloor{\frac{W}{P}}\right \rfloor$ separate patches $x_i, i \in {1, \dots, N}$ of size $P \times P$ each.
The vision encoder we use is a ViT~\cite{dosovitskiy2021image}, which maps the image patches $x_i$ to a vector of $N$ spatial tokens $f(x) = [f(x_{1}), \dots f(x_{N})]$.

\paragraph{\ours\ fine-tuning loss.}
To train the ViT's spatial tokens, we first randomly crop the image $V$-times into $v_g$ global views and $v_l$ local views.
When sampling the views we compute their pairwise intersection in bounding box format and store it in a matrix $B$. 
We denote the transformed version of the image as $x_{t_j}, j\in\{1, \dots, V\}$.
Then, we forward the spatial tokens through a MLP projection head $g$ with a L2-normalization bottleneck to get spatial features for each crop: 
$g(f(x_{t_j})) = Z_{t_j}\ \in \mathbb{R}^{D \times N}$. 
To create prediction targets, we next find an optimal soft cluster assignment $Q_{t_j}$ of all spatial token's feature vector $Z_{t_j}$ to $K$ prototype vectors  $[c_1, \dots, c_K] = C \in \mathbb{R}^{D \times K}$. 
For that, we follow the online optimization objective of SwAV~\cite{caron2020unsupervised} that works on the whole image batch $b$.
$Q$ is optimized such that the similarity between all feature vectors in the batch and the prototypes is maximized, while at the same time being regularized towards assigning equal probability mass to each prototype vector.
This can be cast to an optimal transport problem and is solved efficiently with the Sinkhorn-Knopp algorithm~\cite{asano2020self, cuturi2013sinkhorn}. 
Instead of optimizing over $|b|$ feature vectors, we instead optimize over $N \! \cdot \! |b|$ spatial feature vectors as we have $N$ spatial tokens for each image.
As our batchsizes are small, we utilize a small queue that keeps the past 8192 features, as is done in SwAV.

With the optimal cluster assignment of all image crops' spatial tokens $Q_{t_k} \in \mathbb{R}^{N \times K}$, we formulate a swapped prediction task:
\begin{align}
    L(x_{t_1}, ... , x_{t_V}) = \sum_{j=0}^{v_g} \sum_{i=0}^V \mathbbm{1}_{k\neq j} l(x_{t_i}, x_{t_j}) 
\end{align}
Here, $l$ is the 2D cross entropy between the softmaxed and aligned cluster assignment predictions and the aligned optimal cluster assignments:
\begin{align}
    l(x_{t_i}, x_{t_j}) = H\big[ (s_{\tau}(\alpha_{B_{j,i}}(g(\Phi(x_{t_i}))^T C), \alpha_{B_{i j}}(Q_{t_j})\big],
    \label{eq:2}
\end{align}

where $H$ is cross-entropy and $s_{\tau}$ a softmax scaled by temperature $\tau$.
We use $L$ to jointly minimize the prototypes $C$ as well as the neural networks $f$ and $g$.
$C$ is further L2-normalized after each gradient step such that $Z^T C$ directly computes the cosine similarity between spatial features and prototypes.

Since global crops capture the majority of an image, we solely use these to compute $Q_{t_j}$, as the spatial tokens can attend to global scene information such that the overall prediction target quality improves.
Further, as local crops just cover parts of images and thus also parts of objects, using these produces cluster assignment predictions that effectively enable object-parts-to-object-category reasoning, an important ability for scene understanding.

\paragraph{Alignment.}
In Equation \ref{eq:2} we introduce the alignment operator $\alpha_{B_{ij}}(\cdot)$.
This is necessary because $x_{t_j}$ and $x_{t_i}$ cover different parts of the input image and thus $Q_{t_j}$ and the cluster assignment prediction $Z_{t_i}^T C$ correspond to different image areas. 
To tackle this, $\alpha(\cdot)$ restores the spatial dimensions $\left \lfloor{\frac{H}{P}}\right \rfloor \times \left \lfloor{\frac{W}{P}}\right \rfloor$ and aligns the tensor using the crop intersection bounding boxes $B_{ji}$ and $B_{ij}$ respectively.
In our experiments we use RoI-Align~\cite{he2017mask} that produces features with a fixed and compatible output size.

\paragraph{Foreground-focused clustering.} To focus the clustering on the foreground tokens, we further leverage the ViT's \texttt{CLS} token attention maps $A_{i} \in [0,1]^N$ of each of its attention heads.
To create a foreground clustering mask that can be used during training, we first average the attention heads to one map and apply a Gaussian filter for smoothing.
We then obtain a binary mask $A_b$ by thresholding the map to keep 60\% of the mass following~\cite{caron2021emerging}.
We use $\alpha_{B_{ji}}$ to align the global crop's attention to the intersection with crop $j$.
The resulting mask is then applied as 0-1 weighting to the 2D cross entropy loss, $l \odot A_b$.
Note that we extract the attention maps and spatial tokens with the same forward pass, thus not impacting training speed.

\subsection{Fully unsupervised semantic segmentation}
In this section we describe our method that enables us to do fully unsupervised semantic segmentation.
Its constituent parts work directly in the learned spatial token embedding space and leverage simple K-means clustering.

\subsubsection{Cluster-based Foreground Extraction (CBFE)}
\label{sec:fg-method}
Under the hypothesis that clusters in our learned embedding space correspond to object parts, we should be able to extract foreground objects by assigning each cluster id to foreground object ($\mathrm{fg}$) or background ($\mathrm{bg}$): $\Theta: \{1, \dots, K\} \rightarrow \{\mathrm{fg},\mathrm{bg}\}$.
Thus, at evaluation time, we construct $\Theta$ without supervision, by using ViT's merged attention maps $A_b$ as a noisy foreground hint.
Similar to how we process the attention maps to focus our clustering pretext on foreground, we average the attention heads, apply Gaussian filtering with a 7x7 kernel size and keep $60\%$ of the mass to obtain a binary mask.
Using train data, we rank all clusters by pixel-wise precision with $A_b$ and find a good threshold $c$ for classifying a cluster as foreground.
This gives us $\Theta$ that we apply to the patch-level clustering to get a foreground mask.

\subsubsection{Overclustering with community detection (CD)}
\label{sec:com-method}
As we will see from Table \ref{tab:full-emb-ablate-method}, the segmentation results improve substantially with higher clustering granularities.
However, this is mainly because overclustering draws on label information to group clusters to ground-truth objects and in the limit of providing one cluster for each pixel, it would be equivalent to providing full supervision signals.
Here, we propose a novel overclustering method that requires no additional supervision at all.
\par
The key idea we leverage is that clusters correspond to object parts, and that a set of object parts frequently co-occur together. 
Thus, local co-occurrence of clusters in an image should provide a hint about an object's constituent parts.
Using co-occurrence statistics to categorize objects has been proposed before in \cite{rabinovich2007objects, galleguillos2008object}. 
However, we are the first to work with object parts and no labels and employ a novel network science method to discover objects.
To group the clusters, we construct an undirected and weighted co-occurence network $G = (V, E, w)$, with $v_i, i \in \{1, \dots, K\}$ corresponding to each cluster.
We use a localized co-occurrence variant that regards the 8-neighborhood up to a pixel distance $d$.
Then, we calculate the conditional co-occurrence probability $P(v_j|v_i)$ for clusters $i$ and $j$ over all images $D$.
With the co-occurrence probabilities at hand, we define $w(e_{i,j}) = \min(P(v_j|v_i), P(v_j|v_i))$.
This asymmetric edge weight definition is motivated by the fact that parts need not be mutually predictive:
For instance, a car windshield might co-occur significantly with sky but presence of a sky is not predictive for a car windshield.

To find communities in $G$, we use the common Infomap algorithm~\cite{rosvall2007maps} as it works with weighted graphs and scales linearly with $|E|$.
It works by leveraging an information-theoretic definition of network communities:
Random walks sample information flow in networks and inform the construction a map $\Theta_K$ from nodes to $M$ communities minimizing the expected description length of a random walk. 
With the discrete many-to-one mapping $\Theta_K: V \rightarrow \{1, \dots, M\}$ obtained from Infomap and computed on train data, we merge the clusters of the validation data to the desired number of ground-truth classes and do Hungarian matching \cite{kuhn1955hungarian}.
Note that Hungarian matching does not extract any meaningful label information; it merely makes the evaluation metric permutation-invariant \cite{ji2018invariant}.
\section{Experiments}\label{s:experiments}
In this section, we evaluate the image patch representations learned by \ours.
We first ablate design decisions of our method to find an optimal configuration.
To evaluate whether some datasets are more information-rich for object parts learning than others, we also ablate training on different datasets.
We further test the performance of our dense representations under a transfer learning setup for semantic segmentation.
Furthermore, we show that \ours\ can also be used for fully unsupervised segmentation requiring no label information at all for evaluation.

\subsection{Setup}
\textbf{Evaluation protocols.}
For all experiments, we discard the projection head used during training.
Instead we directly evaluate the ViT's spatial tokens.
We use two main techniques for evaluation: linear classifier and overclustering.
For linear classifier (LC), we fine-tune a 1x1 convolutional layer on top of the frozen spatial token or the pre-GAP \texttt{layer4} features, following \cite{vangansbeke2020unsupervised}.
For overclustering, we run K-Means on all spatial tokens of a given dataset.
We then group cluster to ground-truth classes by greedily matching by  pixel-wise precision and run Hungarian matching \cite{kuhn1955hungarian} on the merged cluster maps to make our evaluation metric permutation-invariant following \cite{ji2018invariant}.
We always report overclustering results averaged over five different seeds.
Overclustering is of special interest as it works directly in the learned embedding space and therefore requires less supervision than training a linear classifier.
For completeness we also report results fine-tuning a deeper fully-convolutional net (FCN) following \cite{wang2021dense}.
Generally, we follow the fine-tuning procedures of prior works \cite{Zhang_2020, vangansbeke2020unsupervised, wang2021dense}.
We report results in mean Intersection over Union (mIoU) unless specified otherwise.

\textbf{Model training.}
We train a ViT-Small with patch size 16 and start training from DINO weights~\cite{caron2021emerging}. 
All models were trained for 50 epochs using batches of size 32 on 2 GPUs.
Further training details are provided in \cref{sec:ap-implementation-details}.

\textbf{Datasets.}
We train our model on ImageNet-100, comprising 100 randomly sampled ImageNet classes \cite{tian2020contrastive}, COCO \cite{lin2014microsoft} and  Pascal VOC (PVOC) \cite{everingham2010pascal}.
When fine-tuning on COCO-Stuff and COCO-Thing, we use a 10\% split of the training sets. 
Evaluation results are computed on the full COCO validation data for COCO-Stuff and COCO-Thing and PVOC12 \textit{val}.
This setup up makes sure that our representations are assessed under varying object definitions (\eg stuff vs. thing) and granularities.
Further details are  provided in the Appendix. 

\subsection{Fine-Tuning Loss Ablations}

In this section, we ablate the most important design decisions and hyperparameters of our fine-tuning loss as well as the aptness of different datasets for learning object parts.
We evaluate on PVOC \textit{val} and report three different overclustering granularities next to LC results.

\begin{table}[]
\setlength{\tabcolsep}{2pt}
\centering
\begin{subtable}[t]{.4\linewidth}
\centering
    \begin{tabular}{llllll}
        \Xhline{2\arrayrulewidth}
        &&  \multicolumn{3}{c}{Num. clusters} \\
        \cmidrule{3-5}
        mask & LC & 100 & 300 & 500 \\ \hline
        all & 67.4 & 37.9 & 44.6 & 47.8  \\
        bg & 64.7 & 28.1 & 39.0 & 41.4\\ 
        fg & \textbf{67.8} & \textbf{38.2} & \textbf{47.2} & \textbf{50.7}\\         \Xhline{2\arrayrulewidth}
    \end{tabular}
\caption{Focusing clustering on foreground (fg) helps.}
\label{tab:mask-ablation}
\end{subtable}%
\hfill
\begin{subtable}[t]{.44\linewidth}
\centering
    \begin{tabular}{lllll}
    \Xhline{2\arrayrulewidth}
    &&  \multicolumn{3}{c}{Num. clusters} \\
            \cmidrule{3-5}
    crops & LC & 100 & 300 & 500 \\ \hline
    {[}2{]} & 66.1 & 33.0 & 42.5 & 45.0 \\
    {[}2,2{]} & 67.7 & 37.8 & 45.4 & 49.3 \\ 
    {[}2,4{]} & \textbf{67.8} & \textbf{38.2} & \textbf{47.2} & \textbf{50.7} \\
    \Xhline{2\arrayrulewidth}
    \end{tabular}
\caption{Local crops boost performance.}
\label{tab:crops-ablation}
\end{subtable}
\\
\vspace{1em}
\begin{subtable}[b]{.45\linewidth}
\centering
    \begin{tabular}{lllll}
    \Xhline{2\arrayrulewidth}
    &&  \multicolumn{3}{c}{Num. clusters} \\
        \cmidrule{3-5}
    tchr & LC & 100 & 300 & 500\\ \hline
    \xmark & 67.6 & 34.6 & 44.3 & 47.9 \\
    \checkmark   & \textbf{67.8} & \textbf{38.2} & \textbf{47.2} & \textbf{50.7} \\
    \Xhline{2\arrayrulewidth}
    \end{tabular}
\caption{Using an EMA teacher helps.}
\label{tab:teacher-ablation}
\end{subtable}
\hfill
\begin{subtable}[b]{.5\linewidth}
\centering
    \begin{tabular}{lllll}
    \Xhline{1.5\arrayrulewidth}
        &&  \multicolumn{3}{c}{Num. clusters} \\
        \cmidrule{3-5}
        protos & LC & 100 & 300 & 500 \\ \hline
        100 & 67.7 & 36.8 & 45.4 & 49.2 \\
        300 & \textbf{67.8} & \textbf{38.2} & \textbf{47.2} & \textbf{50.7} \\
        500 & 67.4 & 35.8 & 44.8 & 49.1 \\ \Xhline{2\arrayrulewidth}
    \end{tabular}
\caption{300 prototypes work well.}
\label{tab:protos-ablation}
\end{subtable}
\vspace{0.5em}
\caption{\textbf{Ablations} of different design decisions for \ours.}
\vspace{0.5em}
\label{tab:full-emb-ablate-method}
\end{table}
\begin{table}[]
\centering
\begin{tabular}{llllllll}
\Xhline{2\arrayrulewidth}
Dataset      & size & LC & K=500 & K=300 & K=100 \\ \hline
IN-100 & 126k & 67.8 & 50.7 & 47.2 & 38.2 \\
COCO         & 118k & \textbf{69.1} & \textbf{53.0} & \textbf{49.9} & \textbf{44.3} \\
PASCAL  & 10k & 64.5 & 50.7 & 47.8 & 38.2 \\ \Xhline{2\arrayrulewidth}
\end{tabular}
\caption{\textbf{Training data study for \ours}. We use the best performing model config from Table \ref{tab:full-emb-ablate-method} and train on different datasets.}
\label{tab:ablate-dataset}
\end{table}

\begin{table*}[t]
    \begin{minipage}[b]{.6\linewidth}
    \setlength{\tabcolsep}{2pt}
      \centering
\begin{tabular}{ll|ll|ll|ll} \Xhline{2\arrayrulewidth}
\multirow{2}{*}{Method} & \multirow{2}{*}{Train} & \multicolumn{2}{c|}{PVOC12} & \multicolumn{2}{c|}{COCO-Things} & \multicolumn{2}{c}{COCO-Stuff} \\
& & LC & K=500 & LC & K=500  & LC & K=500 \\ \hline
\textcolor{mygray}{Sup. ViT} & \textcolor{mygray}{IN + IN21} & \textcolor{mygray}{68.1} & \textcolor{mygray}{55.1} & \textcolor{mygray}{65.2} & \textcolor{mygray}{50.9} & \textcolor{mygray}{49.0} & \textcolor{mygray}{35.1}  \\
\textcolor{mygray}{Sup. ResNet} & \textcolor{mygray}{IN} & \textcolor{mygray}{53.8} & \textcolor{mygray}{36.5} & \textcolor{mygray}{57.8} & \textcolor{mygray}{44.2} & \textcolor{mygray}{44.4} & \textcolor{mygray}{30.8} \\ \hline
\textit{instance-level:} &&&&&&\\
\quad MoCo-v2 \cite{he2019momentum} & IN & 45.0$^{\dagger}$ & 39.1 & 47.5 & 36.2 & 32.6 & 28.3 \\
\quad DINO \cite{caron2021emerging} & IN & 50.6 & 17.4 & 50.6 & 23.5 & 47.7 & 32.1  \\
\quad SwAV \cite{caron2020unsupervised} & IN & 50.7$^{\dagger}$ & 35.7 & 56.7 & 37.3 & 46.0 & 33.1 \\ \hline
\textit{pixel/patch-level:} &&&&&& \\
\quad IIC \cite{ji2018invariant} & PVOC & 28.0$^{\dagger}$ & - & - & - & - & -  \\
\quad  MaskContrast \cite{vangansbeke2020unsupervised} & IN+PVOC & 49.2 & 45.4 & 47.5 & 37.0  & 32.0 & 25.6 \\
\quad DenseCL \cite{wang2021dense} & IN & 49.0 & 43.6 & 53.0 & 41.0 & 40.9 & 30.3 \\\hline
\quad  \textbf{\ours}\ & IN & \underline{68.0} & \underline{50.5} & \underline{62.5} & \underline{49.2} & \underline{51.2} & \textbf{43.8} \\
\quad \textbf{\ours}\ & IN+CC & \textbf{69.3} & \textbf{53.3} & \textbf{67.6} & \textbf{55.9} & \textbf{53.5} & \underline{43.6}  \\ 
\Xhline{2\arrayrulewidth}
\end{tabular}
\caption{\textbf{Transfer learning for semantic segmentation results.} Best results are in \textbf{bold} and second best are \underline{underlined}. 'IN', 'IN21', 'CC' and 'PVOC' indicate training on ImageNet, ImageNet21k, CoCo and Pascal \textit{trainaug} respectively. \mbox{$^\dagger$ indicates} result taken from \cite{vangansbeke2020unsupervised}.}
\label{tab:sota-lin}
    \end{minipage}%
    \hfill%
    \begin{minipage}[b]{.35\linewidth}
    \footnotesize
        \setlength{\tabcolsep}{2pt}
    \centering
\begin{tabular}{ll} \Xhline{2\arrayrulewidth}
Method                 & mIoU \\ \hline
Sup. ResNet & 18.5 \\ 
Sup. ViT & 21.1 \\
\hline
DINO \cite{caron2021emerging} & 4.6 \\
SwAV \cite{he2019momentum}    & 13.7 \\
MoCo-v2 \cite{he2019momentum} & 18.5 \\ 
MaskContrast \cite{vangansbeke2020unsupervised} & 35.0$^{\dagger}$ \\
\hline
\textbf{\ours\ (CBFE+CD)} & \textbf{41.7} \\ \Xhline{2\arrayrulewidth}
\end{tabular}
\caption{\textbf{Unsupervised semantic segmentation results.} 
We outperform other state-of-the-art methods by a large margin. \mbox{$^\dagger$ indicates} result taken from \cite{vangansbeke2020unsupervised}.
}
\label{tab:sota-kmeans-pascal}
      \centering
      \vspace{1em}
       \begin{tabular}{ll} 
        \Xhline{2\arrayrulewidth}
         & mIoU \\ \hline
        K=150 & 48.8 \\ \hline
        DINO   & 4.6 \\ 
        + \ours   & 18.9 (+14.3\%) \\ 
        + CBFE & 36.6 (+17.7\%) \\  
        + CD & 41.7 (+5.1\%) \\ \hline
        \Xhline{2\arrayrulewidth}
        \end{tabular}
    \caption{\textbf{Component contributions.} We show the gains that each individual component brings for PVOC segmentation and K=21.}
    \label{tab:oursvsdino}
    \end{minipage} 
\end{table*}
\paragraph{Model Ablation.}
In Table \ref{tab:full-emb-ablate-method} we report the model ablation results. 
As described in Section \ref{sec:method}, we propose to leverage attention maps to guide our clustering algorithm. 
Note that the attention maps are just a noisy approximation of foreground objects.
Thus, it only \textit{focuses} spatial token clustering on foreground but does not neglect objects in the background.
We find that foreground clustering gives substantial performance gains over our two ablated versions: clustering of all spatial tokens (up to $3\%$) and clustering mostly background tokens (up to $10\%$), as shown in Table \ref{tab:mask-ablation}. 

In Table \ref{tab:crops-ablation} we ablate the multi-crop augmentation strategy.
More specifically, we compare using four local crops against using only two or no local crops.
The usage of local crops (last vs. second to last row) gives a much larger performance gain than using just more local crops (second to last vs. first row).
This shows that predicting cluster assignments from constrained local image information is an important aspect for learning expressive dense representations.
Interestingly, the overclustering results are effected more by this ablation, showing that local crops are important for learning a semantically-structured embedding space.

We also ablate the number of prototypes used for Sinkhorn-Knopp clustering in Table \ref{tab:protos-ablation}. 
We find that the best performance is achieved with a moderate overclustering of 300 prototypes. 
Note however, that the number of prototypes we use for training is not equivalent to the number of clusters used for evaluation, which we denote for instance by K=500.
Lastly, even though we fine-tune a pretrained model, we find that an EMA teacher still helps with learning more expressive representations as can be seen in Table \ref{tab:teacher-ablation}.

\paragraph{Training Data.}
In Table \ref{tab:ablate-dataset} we report results under varying training data: ImageNet-100, COCO and PVOC.
ImageNet-100 usually features object-centric images with few objects.
In comparison, COCO and PVOC contain images with more complex scenes.
For comparability, we adapt the number of epochs for PVOC to 500 such that all models are trained for the same number of iterations.
We find that our method's performance improves by up to $6\%$ when trained on COCO instead of ImageNet-100.
This shows the potential of our dense clustering loss when applied to less object-centric images and is in stark contrast to other methods reporting that their results get worse when training on COCO instead of ImageNet~\cite{Zhang_2020, wang2021dense, li2021dense}.
Finally, we see the worse results on PVOC as a confirmation of the fact that even for fine-tuning, larger datasets perform better for ViTs.

\begin{table}
\centering
\begin{tabular}{ll|ll} \Xhline{2\arrayrulewidth}
\multirow{2}{*}{Method} & \multirow{2}{*}{Train} & \multicolumn{2}{c}{PVOC12} \\
& & FCN \\
\hline
SegSort \cite{hwang2019segsort}     & CC+PVOC & 36.2$^{\dagger}$ \\
Hier. Grouping \cite{zhang2020self} & CC+PVOC & 48.8$^{\dagger}$ \\
DINO \cite{caron2021emerging} & IN & 60.6 \\
Hier. Grouping \cite{zhang2020self} & IN      & 64.7$^{\dagger}$  \\
MoCo-v2 \cite{he2019momentum}       & CC      & 64.5$^{\dagger}$ \\
MoCo-v2 \cite{he2019momentum}       & IN      & 67.5$^{\dagger}$ \\
DenseCL \cite{wang2021dense}        & CC      & 67.5$^{\dagger}$  \\
DenseCL \cite{wang2021dense}        & IN      & 69.4$^{\dagger}$ \\ 
\hline
\textbf{\ours}\                              & IN      & \underline{70.1} \\
\textbf{\ours}\                              & IN+CC   & \textbf{71.4} \\ \hline
\textbf{\ours} (ViT-B/8) & IN+CC & \textbf{76.3} \\
\Xhline{2\arrayrulewidth}
\end{tabular}
\caption{\textbf{FCN transfer learning results}. We follow the same notation as in Table \ref{tab:sota-lin}. Note that Hierarchical Grouping and Segsort fine-tune a larger ASPP decoder. $^\dagger$ indicates result taken from \cite{wang2021dense, zhang2020self}}

\label{tab:sota-fcn}
\end{table}
\begin{figure*}[htp]
     \begin{subfigure}[b]{0.495\columnwidth}
         \centering
         \includegraphics[width=\columnwidth]{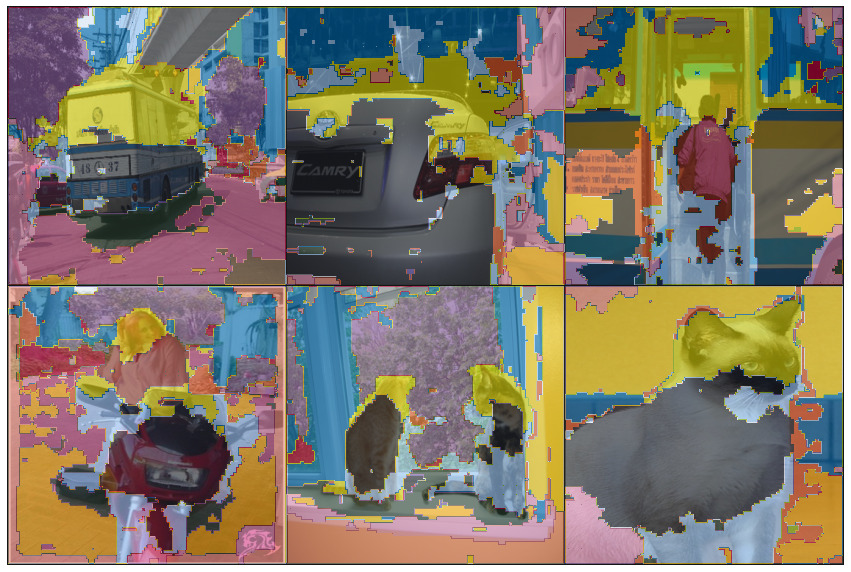}
         \caption{DINO}
         \label{fig:dino-clus}
     \end{subfigure}%
     \hfill
     \begin{subfigure}[b]{0.495\columnwidth}
         \centering
         \includegraphics[width=\columnwidth]{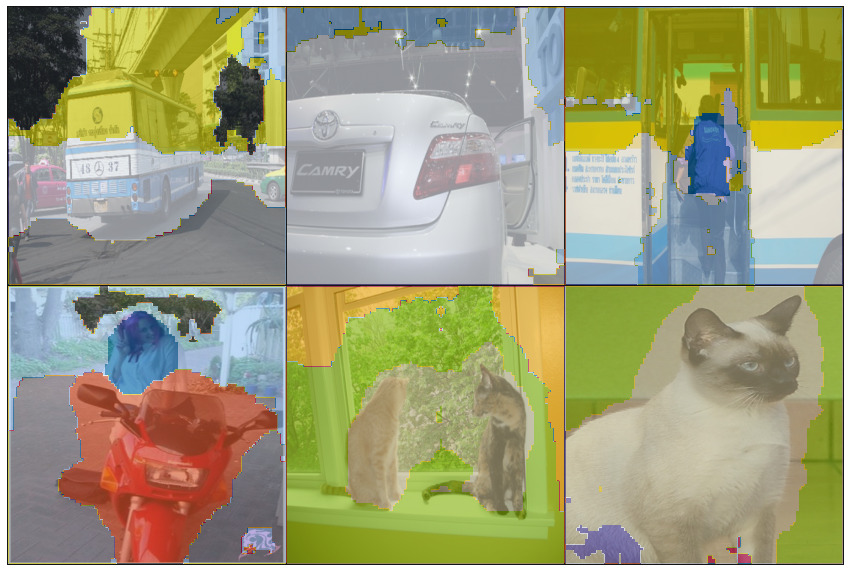}
         \caption{+ \ours}
         \label{fig:our-loss-clus}
     \end{subfigure}%
     \hfill
     \begin{subfigure}[b]{0.495\columnwidth}
         \centering
         \includegraphics[width=\textwidth]{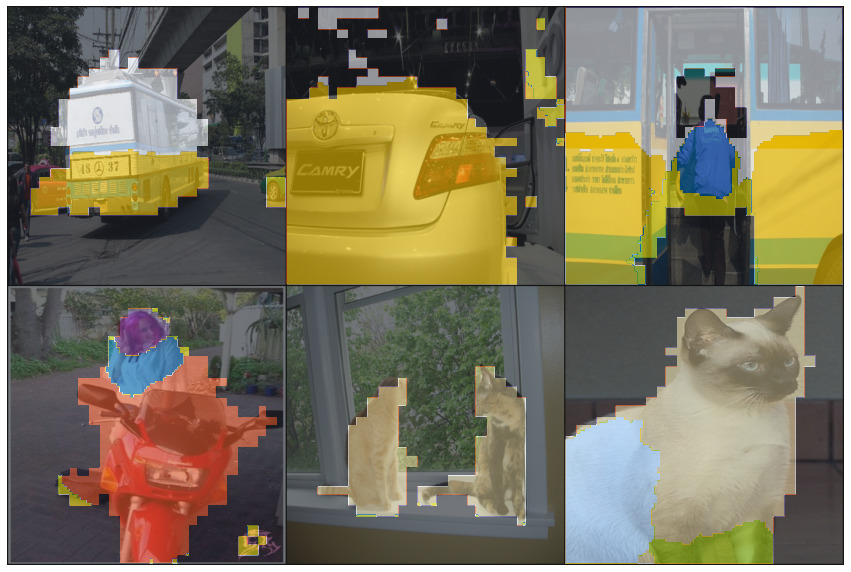}
         \caption{+ CBFE}
         \label{fig:ours-fg-clus}
     \end{subfigure}%
     \hfill
     \begin{subfigure}[b]{0.495\columnwidth}
         \centering
         \includegraphics[width=\textwidth]{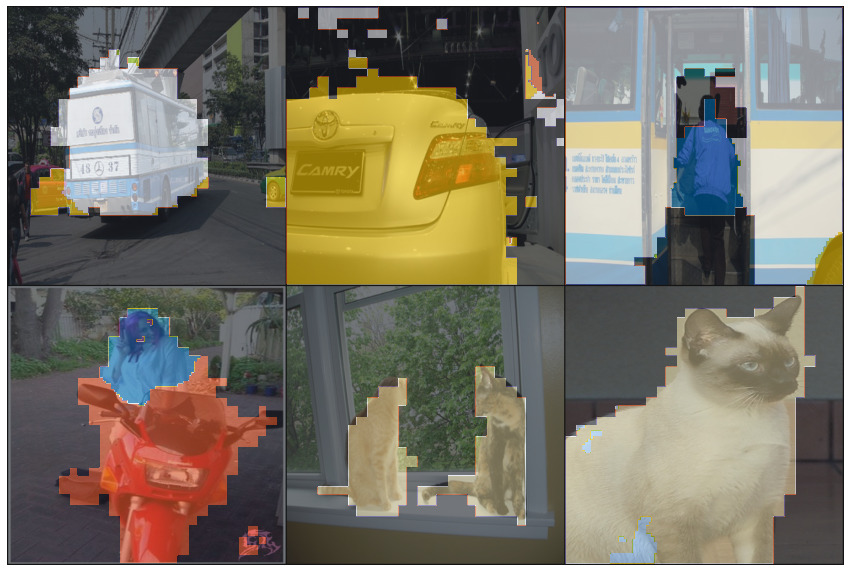}
         \caption{+ CD}
         \label{fig:ours-cd-clus}
     \end{subfigure}
    \caption{\textbf{Qualitative Segmentations by DINO and our gradual  improvements}. We cluster the spatial tokens and visualize the resulting clusters obtained after each step of our method.}
    \label{fig:oursvsdino}
\end{figure*}
\subsection{Transfer learning}
Next, we study how well our dense representations, once learned, generalize to other datasets.
We train our model on ImageNet-100 or COCO and report LC and overclustering results on PVOC12, COCO-Things and COCO-Stuff.
As shown in Table \ref{tab:sota-lin}, we outperform self-supervised prior works by large margins \textit{on all three datasets} even though some use further datasets and supervision.
On PVOC12 we surpass the state-of-the-art by more than $17\%$ for linear evaluation and by more than $5\%$ for overclustering.
On COCO-Things and COCO-Stuff we improve linear classifier by $>5\%$ and $>3\%$ and overclustering by $>8\%$ and $>10\%$ respectively.
Note that these gains are not due to the DINO initialisation nor due to ViTs per-se as the starting DINO model performs on par with other instance-level self-supervised methods that use ResNets like SwAV.
In fact, DINO's embedding space exhibits inferior semantic structure in comparison to MoCo-v2 and SwAV as can be seen from the overclustering results on PVOC12 (-18\%) and COCO-Things (-12\%).
Our method is also on par with the performance of a supervised ViT even though it was trained on a $>$10x times larger full ImageNet (IN-21k) dataset~\cite{ridnik2021imagenet21k}.
When fine-tuning on COCO instead of IN-100, we see further improvements on all datasets.
The results confirm that it is desirable to learn object parts representations, as they work well under different object definitions, as evidenced by strong performances across datasets.

In Table \ref{tab:sota-fcn}, we evaluate \ours\ by fine-tuning a full FCN on top of frozen features.
Again, we outperform all prior works, including DenseCL, the current state-of-the-art.
Interestingly, while DenseCL shows a performance gain of more than $20\%$ when fine-tuning a FCN instead of a linear layer, our performance gain from fine-tuning is relatively low at around $2\%$.
We hypothesize that this behaviour is because our learned embedding space is already close to maximally informative for semantic segmentation under linear transformations.
In contrast, DenseCL's embedding space alone is less informative in itself and requires a more powerful non-linear transformation.
We push state-of-the-art even further by fine-tuning a larger ViT-Base with patch size 8 (ViT-B/8) improving FCN performance by around $5\%$. We report further details and experiments in the Appendix \cref{sec:ap-additional-experiments}.

\subsection{Fully unsupervised semantic segmentation}
\label{sec:full-unsup}
Encouraged by our strong K=500 overclustering results in Table~\ref{tab:sota-lin}, we next evaluate fully unsupervised semantic segmentation.
This relies only on the learned embedding space's structure and refrains from using any test-time label information, \ie the number of final clusters needs to be equivalent to the ground-truth.
To that extent, we start with a simple K-means clustering of the spatial tokens to get cluster assignments for each token.
As prior works, we base our evaluation on PVOC12 \textit{val} and train self-supervised on an arbitrary dataset \cite{vangansbeke2020unsupervised}, in this case COCO.
In Table \ref{tab:sota-kmeans-pascal} we compare our method to prior state-of-the-art.
We outperform our closest competitor, MaskContrast, by $>6\%$.
While like MaskConstrast, we cluster only foreground tokens, we use our embedding space clustering instead of a pretrained unsupervised saliency estimator to do cluster-based foreground extraction (CBFE).
Also, instead of averaging the feature representations per image, we use our novel unsupervised overclustering method with community detection (CD), allowing us to detect multiple object categories in one image.

\subsubsection{Performance gain study}
In Figure \ref{fig:oursvsdino}, we show the gradual visual improvement of the segmentations.
By using \ours\ we substantially improve the DINO baseline by more than $14\%$, as shown quantitatively in Table \ref{tab:oursvsdino}.
This is also apparent when comparing Figure \ref{fig:dino-clus} to Figure \ref{fig:our-loss-clus}.
The DINO segmentations show no correspondence to object categories, whereas the segmentations obtained by \ours\ assign the same colors to the bus in the first and third image of the top row as an example.
However, our segmentations do not correspond well with PVOC's object definitions, as we oversegment background.
To further improve this, we extract foreground resulting in the segmentation maps shown in Figure \ref{fig:ours-fg-clus}.
The segmentation focuses on the foreground and object categories start to emerge more visibly.
However, some objects are still oversegemented such as busses and cats.
Thus, we run our proposed community detection algorithm to do fully unsupervised overclustering, resulting in the segmentations shown in Figure \ref{fig:ours-cd-clus}.

\begin{table}[]
\centering
\begin{tabular}{lll}
\Xhline{2\arrayrulewidth}
Method      & Jacc. (\%)  & B-F1~\cite{csurka2013good} (\%) \\ \hline
DINO attention \cite{caron2021emerging} & 48.7 & 36.5 \\
Unsup. saliency~\cite{vangansbeke2020unsupervised} & 55.9 & \underline{40.8} \\
\ours\ IN CBFE & \underline{58.6} & \textbf{42.1} \\ 
\ours\ CC CBFE & \textbf{59.6} & 40.7 \\ \Xhline{2\arrayrulewidth}
\end{tabular}
\caption{\textbf{Foreground extraction results on PVOC val}. Our method improves over DINO atenttions with respect to Jaccard distance and Boundary F1 score and shows performance on par with a dedicated unsupervised saliency estimator.}
\label{tab:fg-extraction}
\end{table}
\begin{figure}
     \begin{subfigure}[b]{0.495\columnwidth}
         \centering
         \includegraphics[width=\columnwidth]{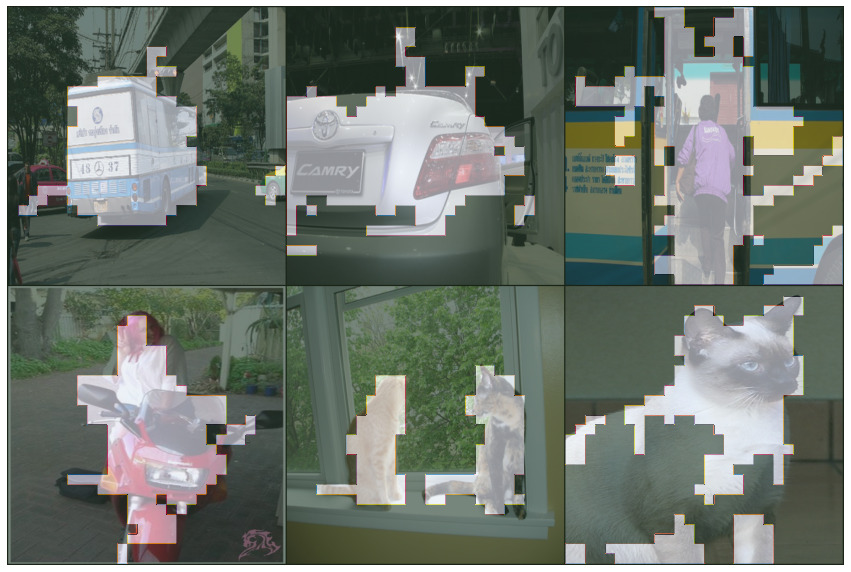}
         \caption{Attention Masks}
     \end{subfigure}%
     \hfill
     \begin{subfigure}[b]{0.495\columnwidth}
         \centering
         \includegraphics[width=\columnwidth]{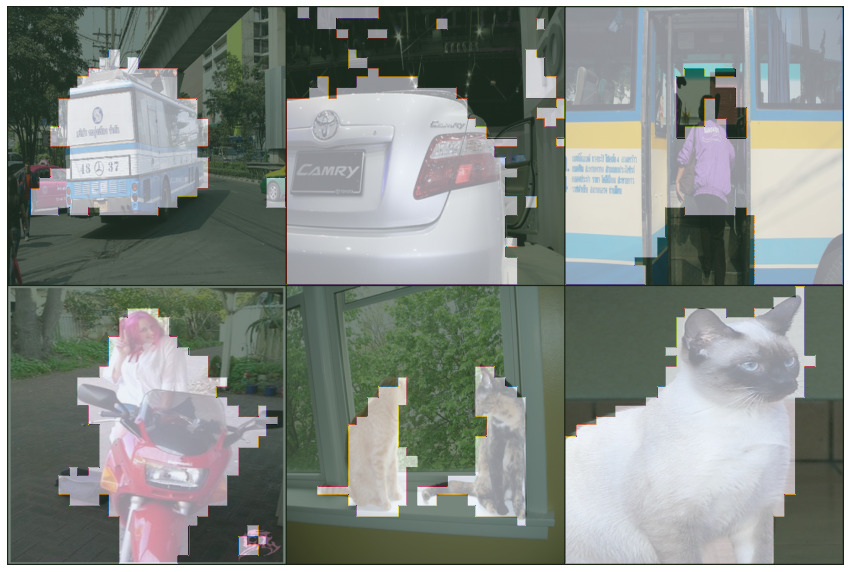}
         \caption{Cluster masks}
     \end{subfigure}%
     \hfill
    \caption{\textbf{DINO Attention masks vs. \ours\ Cluster masks}.}
    \label{fig:masks}
\end{figure}
\paragraph{CBFE.}
\label{sec:fg-extraction}
For foreground extraction, we follow the method proposed in Section \ref{sec:fg-method}.
As shown in Table \ref{tab:fg-extraction}, our foreground masks obtained through CBFE outperform DINO's attention maps by more than 9\%.
This is remarkable as we can only improve the attention map if the foreground clusters also segment the foreground correctly where the noisy foreground hint from DINO's attention is wrong.
In Figure \ref{fig:masks}, we show mask visualizations to provide a qualitative idea of this phenomenon.
While the attention masks only mark the most discriminative regions they fail to capture the foreground object's shape (Fig.~\ref{fig:masks}(a)). 
Our cluster masks, however, alleviate this providing a crisp foreground object segmentation (Fig.~\ref{fig:masks}(b)).
With the foreground masks extracted, we can specify K-Means to run only on foreground spatial tokens.
This further improves our fully unsupervised segmentation performance by $>17\%$, as shown in Table \ref{tab:oursvsdino}.

\begin{figure}
    \centering
    \includegraphics[width=\columnwidth]{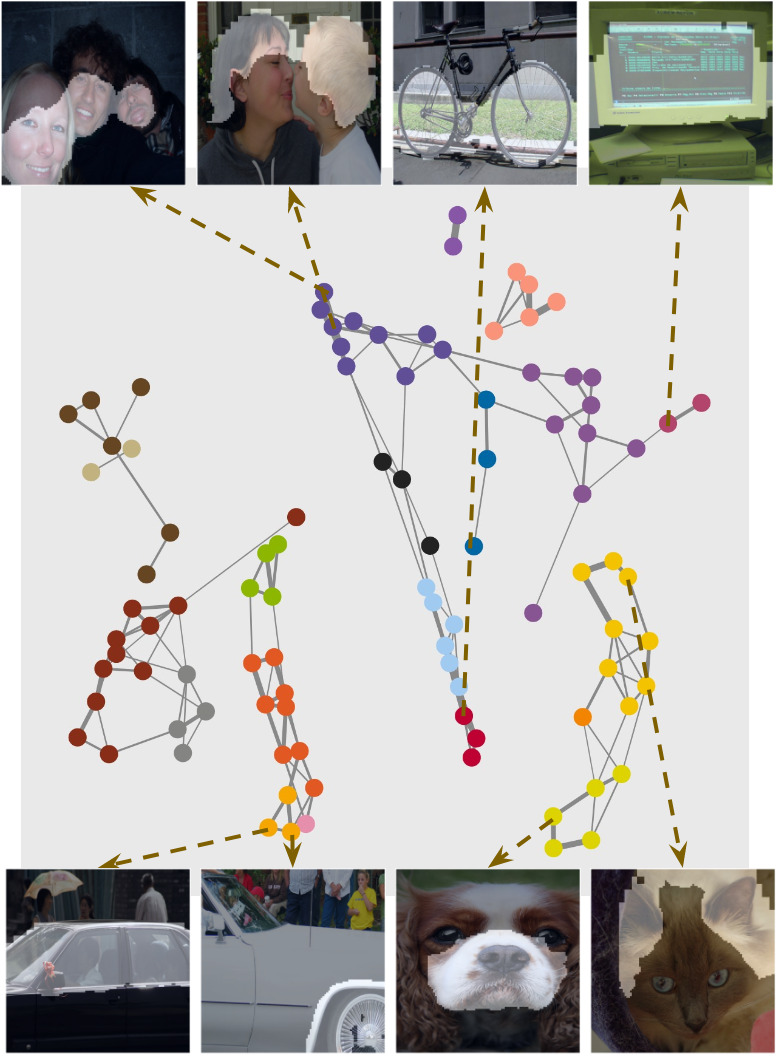}
    \caption{\textbf{Communities found in our cluster co-occurrence network constructed through self-supervision.} Each node corresponds to a cluster in our learnt embedding space. The nodes are colored by community membership.}
    \label{fig:network}
\end{figure}
\paragraph{CD.}
We have seen that overclustering yields benefits in terms of performance but requires additional supervision for merging clusters during evaluation.
To reap the benefits of this process whilst staying unsupervised, we construct a network based on cluster co-occurrences and run community detection (CD) following the method proposed in Section \ref{sec:com-method}.
We find that CD can further improve our performance by $>5\%$ and brings our fully unsupervised semantic segmentation results closer to the upper bound of supervised overclustering at test-time with $K=150$, as shown in Table \ref{tab:oursvsdino}.

Finally, in Figure \ref{fig:network} we show a visualization of the constructed network, the discovered communities as well as some exemplary parts clusters.
For instance, \ours\ discovers bicycle wheels and car wheels separately.
This demonstrates that we can learn high-level semantic clusters that do not latch on low-level information such as shape.
Furthermore, we can observe that clusters that are semantically similar, such as human hair and human faces, are also part of the same community and close in the resulting network. 
Also, a gradual semantic transition within connected components can be observed as shown for dog snout and cat ears being part of different communities that are inter-connected.
\section{Discussion}

\paragraph{Limitations.}
Since we do not learn on a pixel but on a patch level, our segmentation maps are limited in their resolution and detection capabilities.
Thus, our method will fail when fine-grained pixel-level segmentation is required or very small objects covering less than an image patch are supposed to be segmented.
Further, our unsupervised overclustering method does a hard assignment of clusters to communities. 
This has the limitation that object parts which occur in several objects are assigned to the wrong object category when they appear in a specific context.
We show an example of this phenomenon in \cref{fig:k100-clusters} in the Appendix, but leave a solution to future work.

\paragraph{Potential negative societal impact.}
Self-supervised semantic segmentation can scale to large datasets with little to no human labelling effort and extract information from it. 
However, as human input is kept to a minimum, rigorous monitoring of the segmentation results is mandatory, as objects might not be segmented in a way that we are used to or problematic biases in the data might be manifested.
Lack of monitoring could have potential negative impacts in areas such as autonomous driving and virtual reality.

\paragraph{Conclusion.}
In this paper, we propose a dense clustering pretext task for the spatial tokens of a ViT that learns a semantically richer embedding space in contrast to other self-supervised ViTs.
We motivate our pretext task by observing that object definitions are brittle and demonstrate object parts learning as a principled alternative.
Our experiments show that this formulation is favorable as we improve state-of-the-art on PVOC, COCO-Stuff and COCO-Thing semantic segmentation benchmarks featuring different object defintions and granularities.
Finally, our embedding space can also be directly used for fully unsupervised segmentation, showing that objects can be defined as co-occurring object parts.

\subsection*{Acknowledgements}
\noindent A.Z. is thankful for using compute resources while interning at Pina Earth.
Y.M.A is thankful for MLRA funding from AWS. 

{\small
\bibliographystyle{ieee_fullname}
\bibliography{bib}

\begin{thebibliography}{10}\itemsep=-1pt

\bibitem{asano2020self}
Yuki~M Asano, Christian Rupprecht, and Andrea Vedaldi.
\newblock Self-labelling via simultaneous clustering and representation
  learning.
\newblock In {\em ICLR}, 2020.

\bibitem{bojanowski2017unsupervised}
Piotr Bojanowski and Armand Joulin.
\newblock Unsupervised learning by predicting noise, 2017.

\bibitem{caesar2018coco}
Holger Caesar, Jasper Uijlings, and Vittorio Ferrari.
\newblock Coco-stuff: Thing and stuff classes in context.
\newblock In {\em Proceedings of the IEEE conference on computer vision and
  pattern recognition}, pages 1209--1218, 2018.

\bibitem{caron2018deep}
Mathilde Caron, Piotr Bojanowski, Armand Joulin, and Matthijs Douze.
\newblock Deep clustering for unsupervised learning of visual features.
\newblock In {\em ECCV}, 2018.

\bibitem{caron2020unsupervised}
Mathilde Caron, Ishan Misra, Julien Mairal, Priya Goyal, Piotr Bojanowski, and
  Armand Joulin.
\newblock Unsupervised learning of visual features by contrasting cluster
  assignments.
\newblock In {\em NeurIPS}, 2020.

\bibitem{caron2021emerging}
Mathilde Caron, Hugo Touvron, Ishan Misra, Herv\'e J\'egou, Julien Mairal,
  Piotr Bojanowski, and Armand Joulin.
\newblock Emerging properties in self-supervised vision transformers.
\newblock In {\em ICCV}, 2021.

\bibitem{chen2020generative}
Mark Chen, Alec Radford, Rewon Child, Jeff Wu, and Heewoo Jun.
\newblock Generative pretraining from pixels.
\newblock In {\em ICML}, 2020.

\bibitem{chen2021empirical}
Xinlei Chen, Saining Xie, and Kaiming He.
\newblock An empirical study of training self-supervised vision transformers.
\newblock {\em arXiv preprint arXiv:2104.02057}, 2021.

\bibitem{cho2015unsupervised}
Minsu Cho, Suha Kwak, Ivan Laptev, Cordelia Schmid, and Jean Ponce.
\newblock Unsupervised object discovery and localization in images and videos.
\newblock In {\em International conference on ubiquitous robots and ambient
  intelligence (URAI)}, pages 292--293. IEEE, 2015.

\bibitem{choudhury21unsupervised}
Subhabrata Choudhury, Iro Laina, Christian Rupprecht, and Andrea Vedaldi.
\newblock Unsupervised part discovery from contrastive reconstruction.
\newblock In {\em NeurIPS}, volume~35, 2021.

\bibitem{chu2021twins}
Xiangxiang Chu, Zhi Tian, Yuqing Wang, Bo Zhang, Haibing Ren, Xiaolin Wei,
  Huaxia Xia, and Chunhua Shen.
\newblock Twins: Revisiting spatial attention design in vision transformers.
\newblock {\em arXiv preprint arXiv:2104.13840}, 2021.

\bibitem{Cordts2016Cityscapes}
Marius Cordts, Mohamed Omran, Sebastian Ramos, Timo Rehfeld, Markus Enzweiler,
  Rodrigo Benenson, Uwe Franke, Stefan Roth, and Bernt Schiele.
\newblock The cityscapes dataset for semantic urban scene understanding.
\newblock In {\em CVPR}, 2016.

\bibitem{csurka2013good}
Gabriela Csurka, Diane Larlus, Florent Perronnin, and France Meylan.
\newblock What is a good evaluation measure for semantic segmentation?.
\newblock In {\em Bmvc}, volume~27, pages 10--5244. Bristol, 2013.

\bibitem{cuturi2013sinkhorn}
Marco Cuturi.
\newblock Sinkhorn distances: Lightspeed computation of optimal transport.
\newblock {\em NeurIPS}, 26:2292--2300, 2013.

\bibitem{edlerMapequation}
A.~Eriksson D.~Edler and M. Rosvall.
\newblock The mapequation software package.
\newblock {\em GitHub. available online at http://www.mapequation.org and
  https://github.com/mapequation/infomap}.

\bibitem{Doersch_2015_ICCV}
Carl Doersch, Abhinav Gupta, and Alexei~A. Efros.
\newblock Unsupervised visual representation learning by context prediction.
\newblock In {\em Proceedings of the IEEE International Conference on Computer
  Vision (ICCV)}, December 2015.

\bibitem{dosovitskiy2021image}
Alexey Dosovitskiy, Lucas Beyer, Alexander Kolesnikov, Dirk Weissenborn,
  Xiaohua Zhai, Thomas Unterthiner, Mostafa Dehghani, Matthias Minderer, Georg
  Heigold, Sylvain Gelly, Jakob Uszkoreit, and Neil Houlsby.
\newblock An image is worth 16x16 words: Transformers for image recognition at
  scale, 2021.

\bibitem{dosovitskiy2015discriminative}
Alexey Dosovitskiy, Philipp Fischer, Jost~Tobias Springenberg, Martin
  Riedmiller, and Thomas Brox.
\newblock Discriminative unsupervised feature learning with exemplar
  convolutional neural networks, 2015.

\bibitem{mae}
He et al.
\newblock Masked autoencoders are scalable vision learners.
\newblock {\em arXiv:2111.06377}, 2021.

\bibitem{chen2021mocov3}
Xinlei~Chen et al.
\newblock An empirical study of training self-supervised vision transformers.
\newblock {\em ICCV}, 2021.

\bibitem{everingham2010pascal}
Mark Everingham, Luc Van~Gool, Christopher~KI Williams, John Winn, and Andrew
  Zisserman.
\newblock The pascal visual object classes (voc) challenge.
\newblock {\em International journal of computer vision}, 88(2):303--338, 2010.

\bibitem{falcon2019pytorch}
et~al. Falcon, WA.
\newblock Pytorch lightning.
\newblock {\em GitHub. Note:
  https://github.com/PyTorchLightning/pytorch-lightning}, 3, 2019.

\bibitem{galleguillos2008object}
Carolina Galleguillos, Andrew Rabinovich, and Serge Belongie.
\newblock Object categorization using co-occurrence, location and appearance.
\newblock In {\em CVPR}, pages 1--8. IEEE, 2008.

\bibitem{gao2021largescale}
Shang-Hua Gao, Zhong-Yu Li, Ming-Hsuan Yang, Ming-Ming Cheng, Junwei Han, and
  Philip Torr.
\newblock Large-scale unsupervised semantic segmentation, 2021.

\bibitem{gidaris2021obow}
Spyros Gidaris, Andrei Bursuc, Gilles Puy, Nikos Komodakis, Matthieu Cord, and
  Patrick Perez.
\newblock Obow: Online bag-of-visual-words generation for self-supervised
  learning.
\newblock In {\em CVPR}, pages 6830--6840, 2021.

\bibitem{gidaris2018unsupervised}
Spyros Gidaris, Praveer Singh, and Nikos Komodakis.
\newblock Unsupervised representation learning by predicting image rotations.
\newblock {\em ICLR}, 2018.

\bibitem{grill2020bootstrap}
Jean-Bastien Grill, Florian Strub, Florent Altch{\'e}, Corentin Tallec,
  Pierre~H Richemond, Elena Buchatskaya, Carl Doersch, Bernardo~Avila Pires,
  Zhaohan~Daniel Guo, Mohammad~Gheshlaghi Azar, et~al.
\newblock Bootstrap your own latent: A new approach to self-supervised
  learning.
\newblock {\em NeurIPS}, 2020.

\bibitem{Hadsell2006DimensionalityRB}
Raia Hadsell, Sumit Chopra, and Yann LeCun.
\newblock Dimensionality reduction by learning an invariant mapping.
\newblock In {\em CVPR}, 2006.

\bibitem{he2019momentum}
Kaiming He, Haoqi Fan, Yuxin Wu, Saining Xie, and Ross Girshick.
\newblock Momentum contrast for unsupervised visual representation learning.
\newblock In {\em CVPR}, 2020.

\bibitem{he2018rethinking}
Kaiming He, Ross Girshick, and Piotr Dollár.
\newblock Rethinking imagenet pre-training, 2018.

\bibitem{he2017mask}
Kaiming He, Georgia Gkioxari, Piotr Doll{\'a}r, and Ross Girshick.
\newblock Mask r-cnn.
\newblock In {\em ICCV}, pages 2961--2969, 2017.

\bibitem{henaff2021efficient}
Olivier~J H{\'e}naff, Skanda Koppula, Jean-Baptiste Alayrac, Aaron van~den
  Oord, Oriol Vinyals, and Jo{\~a}o Carreira.
\newblock Efficient visual pretraining with contrastive detection.
\newblock {\em arXiv preprint arXiv:2103.10957}, 2021.

\bibitem{hendrycks2016gaussian}
Dan Hendrycks and Kevin Gimpel.
\newblock Gaussian error linear units (gelus).
\newblock {\em arXiv preprint arXiv:1606.08415}, 2016.

\bibitem{hung2019scops}
Wei-Chih Hung, Varun Jampani, Sifei Liu, Pavlo Molchanov, Ming-Hsuan Yang, and
  Jan Kautz.
\newblock Scops: Self-supervised co-part segmentation.
\newblock In {\em CVPR}, pages 869--878, 2019.

\bibitem{hwang2019segsort}
Jyh-Jing Hwang, Stella~X Yu, Jianbo Shi, Maxwell~D Collins, Tien-Ju Yang, Xiao
  Zhang, and Liang-Chieh Chen.
\newblock Segsort: Segmentation by discriminative sorting of segments.
\newblock In {\em CVPR}, pages 7334--7344, 2019.

\bibitem{jenni2018selfsupervised}
Simon Jenni and Paolo Favaro.
\newblock Self-supervised feature learning by learning to spot artifacts, 2018.

\bibitem{ji2018invariant}
Xu Ji, João~F. Henriques, and Andrea Vedaldi.
\newblock Invariant information clustering for unsupervised image
  classification and segmentation.
\newblock In {\em ICCV}, 2019.

\bibitem{JDH17}
Jeff Johnson, Matthijs Douze, and Herv{\'e} J{\'e}gou.
\newblock Billion-scale similarity search with gpus.
\newblock {\em arXiv preprint arXiv:1702.08734}, 2017.

\bibitem{kirillov2019panoptic}
Alexander Kirillov, Kaiming He, Ross Girshick, Carsten Rother, and Piotr
  Doll{\'a}r.
\newblock Panoptic segmentation.
\newblock In {\em CVPR}, pages 9404--9413, 2019.

\bibitem{kuhn1955hungarian}
Harold~W Kuhn.
\newblock The hungarian method for the assignment problem.
\newblock {\em Naval research logistics quarterly}, 2(1-2):83--97, 1955.

\bibitem{li2021dense}
Xiaoni Li, Yu Zhou, Yifei Zhang, Aoting Zhang, Wei Wang, Ning Jiang, Haiying
  Wu, and Weiping Wang.
\newblock Dense semantic contrast for self-supervised visual representation
  learning.
\newblock {\em arXiv preprint arXiv:2109.07756}, 2021.

\bibitem{lin2014microsoft}
Tsung-Yi Lin, Michael Maire, Serge Belongie, James Hays, Pietro Perona, Deva
  Ramanan, Piotr Doll{\'a}r, and C~Lawrence Zitnick.
\newblock Microsoft coco: Common objects in context.
\newblock In {\em ECCV}, pages 740--755, 2014.

\bibitem{liu2021selfemd}
Songtao Liu, Zeming Li, and Jian Sun.
\newblock Self-emd: Self-supervised object detection without imagenet, 2021.

\bibitem{liu2021swin}
Ze Liu, Yutong Lin, Yue Cao, Han Hu, Yixuan Wei, Zheng Zhang, Stephen Lin, and
  Baining Guo.
\newblock Swin transformer: Hierarchical vision transformer using shifted
  windows.
\newblock {\em ICCV}, 2021.

\bibitem{noroozi2016unsupervised}
Mehdi Noroozi and Paolo Favaro.
\newblock Unsupervised learning of visual representations by solving jigsaw
  puzzles.
\newblock In {\em ECCV}, 2016.

\bibitem{NEURIPS2019_9015}
Adam Paszke, Sam Gross, Francisco Massa, Adam Lerer, James Bradbury, Gregory
  Chanan, Trevor Killeen, Zeming Lin, Natalia Gimelshein, Luca Antiga, Alban
  Desmaison, Andreas Kopf, Edward Yang, Zachary DeVito, Martin Raison, Alykhan
  Tejani, Sasank Chilamkurthy, Benoit Steiner, Lu Fang, Junjie Bai, and Soumith
  Chintala.
\newblock Pytorch: An imperative style, high-performance deep learning library.
\newblock In H. Wallach, H. Larochelle, A. Beygelzimer, F. d\textquotesingle
  Alch\'{e}-Buc, E. Fox, and R. Garnett, editors, {\em Advances in Neural
  Information Processing Systems 32}, pages 8024--8035. 2019.

\bibitem{pathak2016context}
Deepak Pathak, Philipp Krahenbuhl, Jeff Donahue, Trevor Darrell, and Alexei~A.
  Efros.
\newblock Context encoders: Feature learning by inpainting, 2016.

\bibitem{peirceobject}
Charles Peirce.
\newblock Reflections on real and unreal objects.

\bibitem{purushwalkam2020demystifying}
Senthil Purushwalkam and Abhinav Gupta.
\newblock Demystifying contrastive self-supervised learning: Invariances,
  augmentations and dataset biases, 2020.

\bibitem{rabinovich2007objects}
Andrew Rabinovich, Andrea Vedaldi, Carolina Galleguillos, Eric Wiewiora, and
  Serge Belongie.
\newblock Objects in context.
\newblock In {\em 2007 IEEE 11th International Conference on Computer Vision},
  pages 1--8. IEEE, 2007.

\bibitem{ridnik2021imagenet21k}
Tal Ridnik, Emanuel Ben-Baruch, Asaf Noy, and Lihi Zelnik-Manor.
\newblock Imagenet-21k pretraining for the masses.
\newblock {\em NeurIPS-Data}, 2021.

\bibitem{roh2021spatially}
Byungseok Roh, Wuhyun Shin, Ildoo Kim, and Sungwoong Kim.
\newblock Spatially consistent representation learning.
\newblock In {\em CVPR}, pages 1144--1153, 2021.

\bibitem{rosvall2007maps}
Martin Rosvall and Carl~T Bergstrom.
\newblock Maps of information flow reveal community structure in complex
  networks.
\newblock {\em arXiv preprint:0707.0609}, 2007.

\bibitem{simeoni2021localizing}
Oriane Sim{\'e}oni, Gilles Puy, Huy~V Vo, Simon Roburin, Spyros Gidaris, Andrei
  Bursuc, Patrick P{\'e}rez, Renaud Marlet, and Jean Ponce.
\newblock Localizing objects with self-supervised transformers and no labels.
\newblock {\em arXiv preprint arXiv:2109.14279}, 2021.

\bibitem{tian2020contrastive}
Yonglong Tian, Dilip Krishnan, and Phillip Isola.
\newblock Contrastive multiview coding.
\newblock In {\em ECCV}, 2020.

\bibitem{van2020scan}
Wouter Van~Gansbeke, Simon Vandenhende, Stamatios Georgoulis, Marc Proesmans,
  and Luc Van~Gool.
\newblock Scan: Learning to classify images without labels.
\newblock In {\em ECCV}, pages 268--285, 2020.

\bibitem{vangansbeke2020unsupervised}
Wouter Van~Gansbeke, Simon Vandenhende, Stamatios Georgoulis, and Luc Van~Gool.
\newblock Unsupervised semantic segmentation by contrasting object mask
  proposals.
\newblock In {\em ICCV}, 2021.

\bibitem{wang2021pyramid}
Wenhai Wang, Enze Xie, Xiang Li, Deng-Ping Fan, Kaitao Song, Ding Liang, Tong
  Lu, Ping Luo, and Ling Shao.
\newblock Pyramid vision transformer: A versatile backbone for dense prediction
  without convolutions.
\newblock {\em ICCV}, 2021.

\bibitem{wang2021dense}
Xinlong Wang, Rufeng Zhang, Chunhua Shen, Tao Kong, and Lei Li.
\newblock Dense contrastive learning for self-supervised visual pre-training.
\newblock In {\em CVPR}, 2021.

\bibitem{wu2021cvt}
Haiping Wu, Bin Xiao, Noel Codella, Mengchen Liu, Xiyang Dai, Lu Yuan, and Lei
  Zhang.
\newblock Cvt: Introducing convolutions to vision transformers.
\newblock {\em ICCV}, 2021.

\bibitem{Wu_2018_CVPR}
Zhirong Wu, Yuanjun Xiong, Stella~X. Yu, and Dahua Lin.
\newblock Unsupervised feature learning via non-parametric instance
  discrimination.
\newblock In {\em CVPR}, 2018.

\bibitem{zhang2019aet}
Liheng Zhang, Guo-Jun Qi, Liqiang Wang, and Jiebo Luo.
\newblock Aet vs. aed: Unsupervised representation learning by auto-encoding
  transformations rather than data, 2019.

\bibitem{zhang2016colorful}
Richard Zhang, Phillip Isola, and Alexei~A Efros.
\newblock Colorful image colorization.
\newblock In {\em ECCV}, 2016.

\bibitem{zhang2020self}
Xiao Zhang and Michael Maire.
\newblock Self-supervised visual representation learning from hierarchical
  grouping.
\newblock {\em arXiv preprint arXiv:2012.03044}, 2020.

\bibitem{Zhang_2020}
Ziqi Zhang, Yaya Shi, Chunfeng Yuan, Bing Li, Peijin Wang, Weiming Hu, and
  Zheng-Jun Zha.
\newblock Object relational graph with teacher-recommended learning for video
  captioning.
\newblock In {\em CVPR}, 2020.

\bibitem{zhou2017scene}
Bolei Zhou, Hang Zhao, Xavier Puig, Sanja Fidler, Adela Barriuso, and Antonio
  Torralba.
\newblock Scene parsing through ade20k dataset.
\newblock In {\em CVPR}, pages 633--641, 2017.

\end{thebibliography}
}
\clearpage
\newpage
\appendix
\section{Appendix}
\begin{figure}[b]
     \begin{subfigure}[b]{0.495\columnwidth}
         \centering
         \includegraphics[width=\columnwidth]{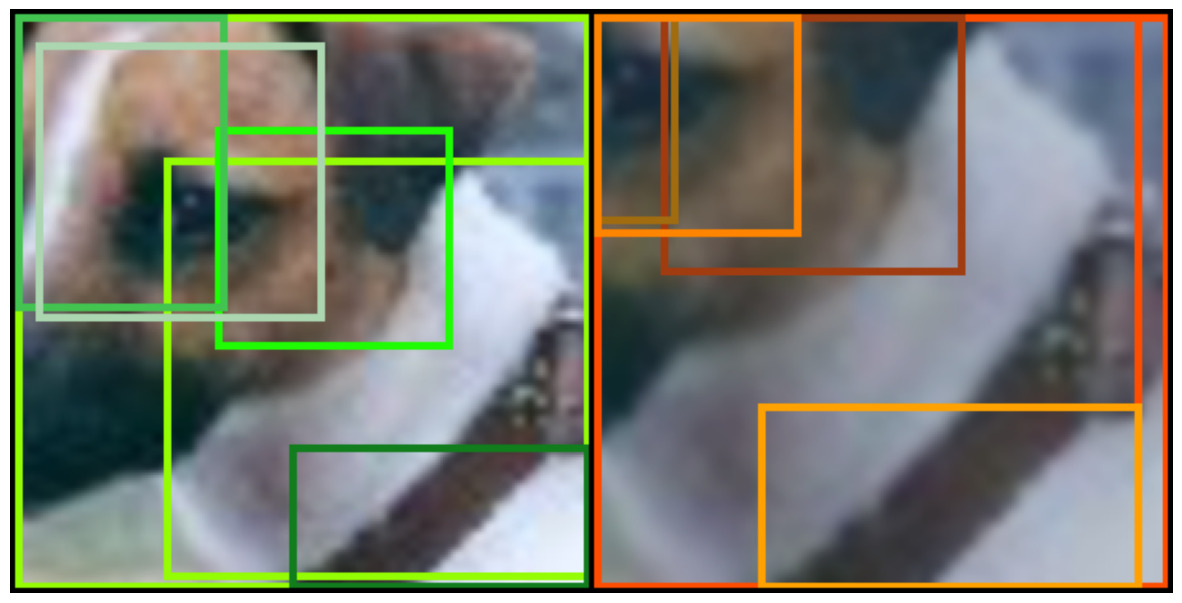}
     \end{subfigure}%
     \begin{subfigure}[b]{0.495\columnwidth}
         \centering
         \includegraphics[width=\columnwidth]{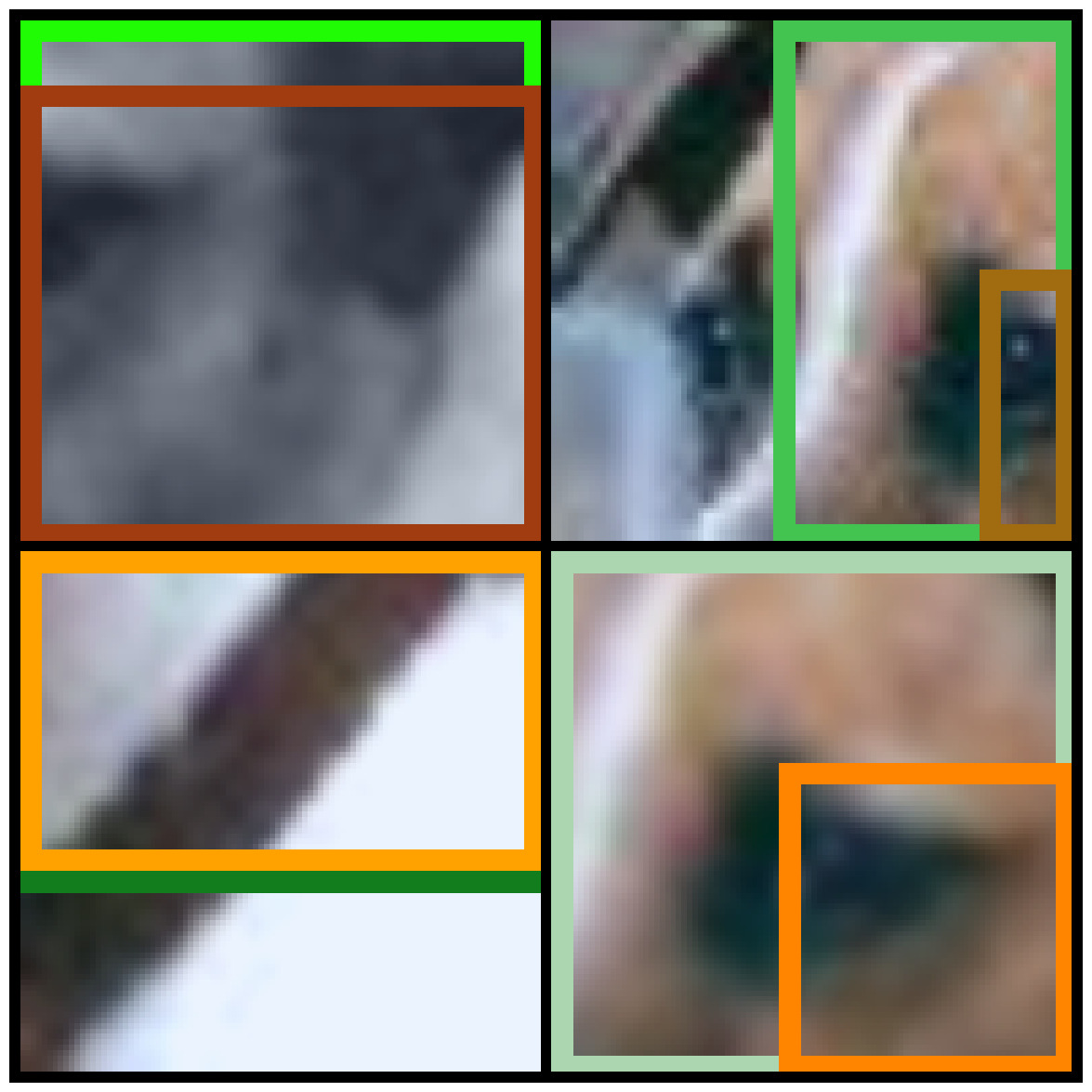}
     \end{subfigure}%
     \hfill
     \begin{subfigure}[b]{0.495\columnwidth}
         \centering
         \includegraphics[width=\columnwidth]{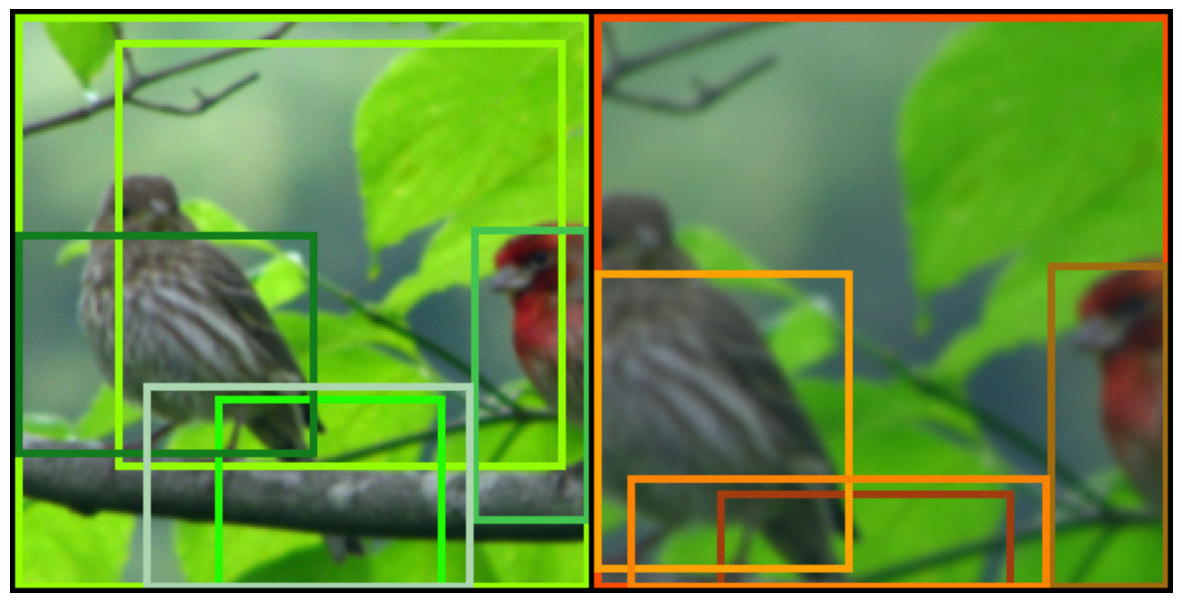}
     \end{subfigure}%
     \begin{subfigure}[b]{0.495\columnwidth}
         \centering
         \includegraphics[width=\columnwidth]{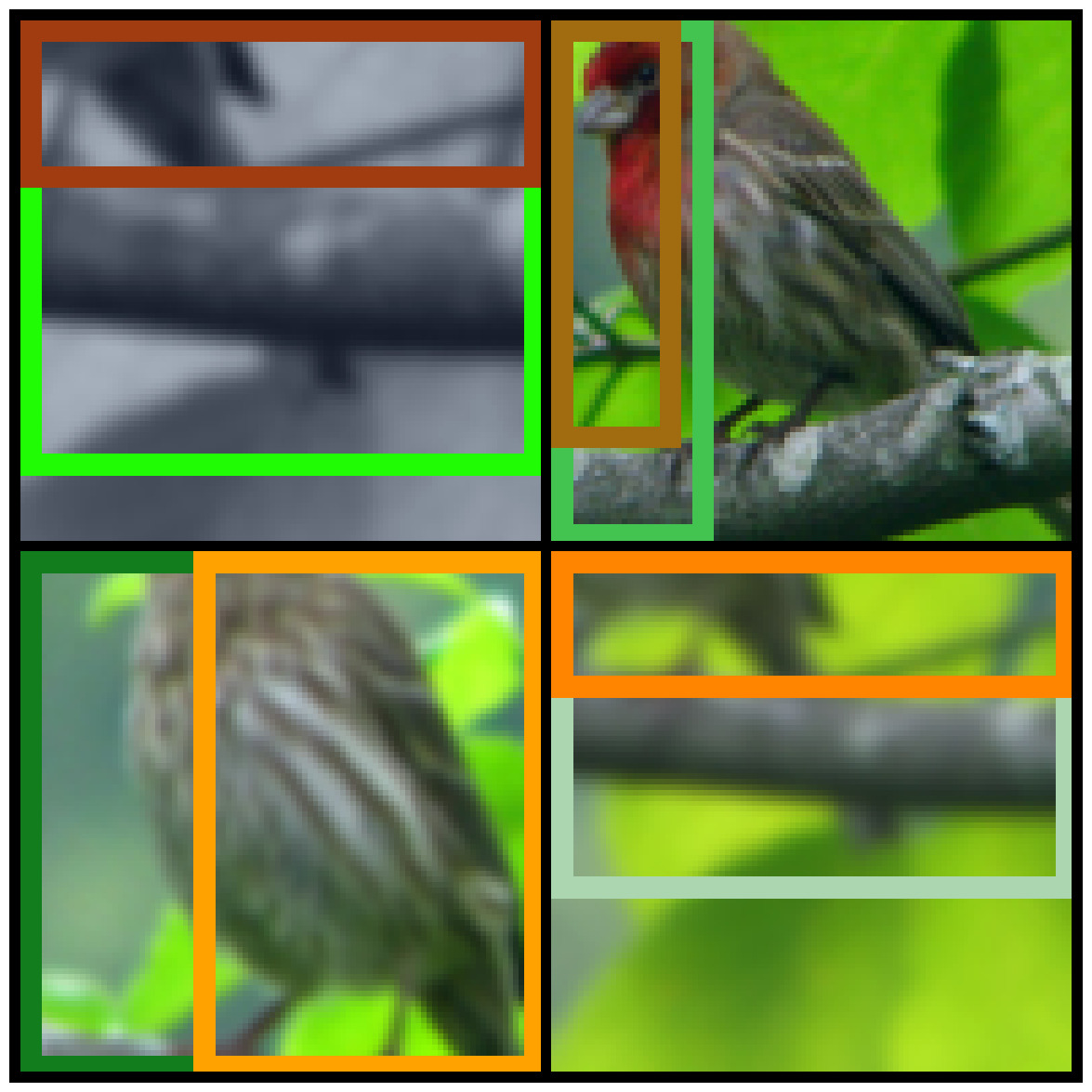}
     \end{subfigure}%
    \caption{\textbf{Bounding box generation example for cluster assignment alginment}. The left column shows the global crops, the right column the local crops. Each global crop has $N-1$ bounding boxes as it produces prediction targets for all remaining $N-1$ crops. Each local crop has $N_{gc}$, the number of global crops, bounding boxes as it is used to predict the prediction targets of each global crop.}
    \label{fig:bbox-generation}
\end{figure}

\subsection{Implementation details}
\label{sec:ap-implementation-details}

\paragraph{Model training}
Our model is implemented in Torch \cite{NEURIPS2019_9015} and PyTorch Lightning \cite{falcon2019pytorch}. 
We use Faiss \cite{JDH17} for K-Means clustering and the MapEquation software package \cite{edlerMapequation} for community detection. 

We chose to train a ViT-Small~\cite{dosovitskiy2021image} as the amount of parameters is roughly equivalent to a ResNet-50's (21M vs. 23M).
Further, we use a student-teacher setup and the teacher weights are updated by the exponential moving average of the student weights following \cite{caron2021emerging, grill2020bootstrap}.
The exponential moving average for updating the teacher weights is adapted with a cosine schedule starting at 0.9995 and going up to 1 i.e. a hard copy.
We train the ViT-Small with a cosine learning rate schedule going down to 0 over  50 training epochs.
The initial projection head learning rate is $1\mathrm{e}{-4}$ and the backbone's learning rate is $1\mathrm{e}{-5}$.
The projection head consists out of three linear layers with hidden dimensionality of 2048 and Gaussian error linear units as activation function \cite{hendrycks2016gaussian}.
The output dimensionality is 256 and the resulting tensors are then passed through a l2-bottleneck and the prototype matrix $C$ to produce cluster assignment predictions.
As discussed, we use a queue for Sinkhorn-Knopp clustering with a length of 8192.
We set the temperature to 0.1 and use Adam as an optimizer with a cosine weight decay schedule.
The alignment happens to a fixed output size of 7x7 during training.
This makes sure that the local and global crop feature maps have the same spatial resolution.
The augmentations used are: random color-jitter, Gaussian blur, grayscale and multi-crop augmentations and the global crop's resolution is 224x224 and the local crop's resolution is 96x96, as in~\cite{caron2020unsupervised}.
We generate global and local crops with the constraint that they intersect at least by $1\%$ of the original image size to make sure that there is a non-negligible intersection where we can apply our clustering loss to.  
In Figure~\ref{fig:bbox-generation} we show the generation process for two ImageNet pictures.

\paragraph{Fully unsupervised semantic segmentation}
For CBFE and CD we take PVOC12 \textit{train} to find good hyperparameter configurations, \ie clustering granularities $K$, the precision threshold for CBFE as well as Markov time and the co-occurrence probability threshold for CD.
We use a segmentation of our embedding space to 200 clusters as we found this granularity to work best on PVOC for CBFE.
Before doing foreground-focused clustering using the cluster mask, we bilinearly interpolate the embeddings to the desired mask size.
For CBFE we report the precision thresholds used in Table \ref{tab:fg_extraction_precision}.
To evaluate the unsupervised saliency estimator baseline method, we use the saliency head provided by the MaskContrast authors \cite{vangansbeke2020unsupervised}.
For the computation of the Jaccard distance we include unlabelled objects as foreground. We can identify these objects as they have a separate class in the PVOC dataset.
\par
For the CD experiments on PVOC, we cluster the embedding space to 150 clusters as we found this granularity to work well here.
To construct the co-occurrence network, we calculate the conditional co-occurrence probability on each image and then average over all images the cluster appeared.
The MapEquation software package can be instructed to constrain the number of found communities.
We use this setting to find exactly as many communities as there are object categories in the given dataset (for PVOC it is 20).
All clusters that are not in communities are assigned to background, which are just 4 out of 100 for our network, as we already focus clustering on foreground.
We set the co-occurrence probability threshold to $9\%$ and all edges below this threshold are ignored by Infomap.
Further as stopping criterion we set the Markov time to $2$ and all other parameters are left at default value.
We report results averaged over 10 seeds.

\begin{table}[b]
\centering
\begin{tabular}{l|l} \Xhline{2\arrayrulewidth}
Method & Precision Threshold \\\hline
\ours\ IN & 35\% \\
\ours\ IN+CC & 40\% \\ \Xhline{2\arrayrulewidth}
\end{tabular}
\caption{\textbf{Precision values used for classifying clusters as foreground.}}
\label{tab:fg_extraction_precision}
\end{table}
\begin{figure*}[htp]
    \centering
    \includegraphics[width=\textwidth]{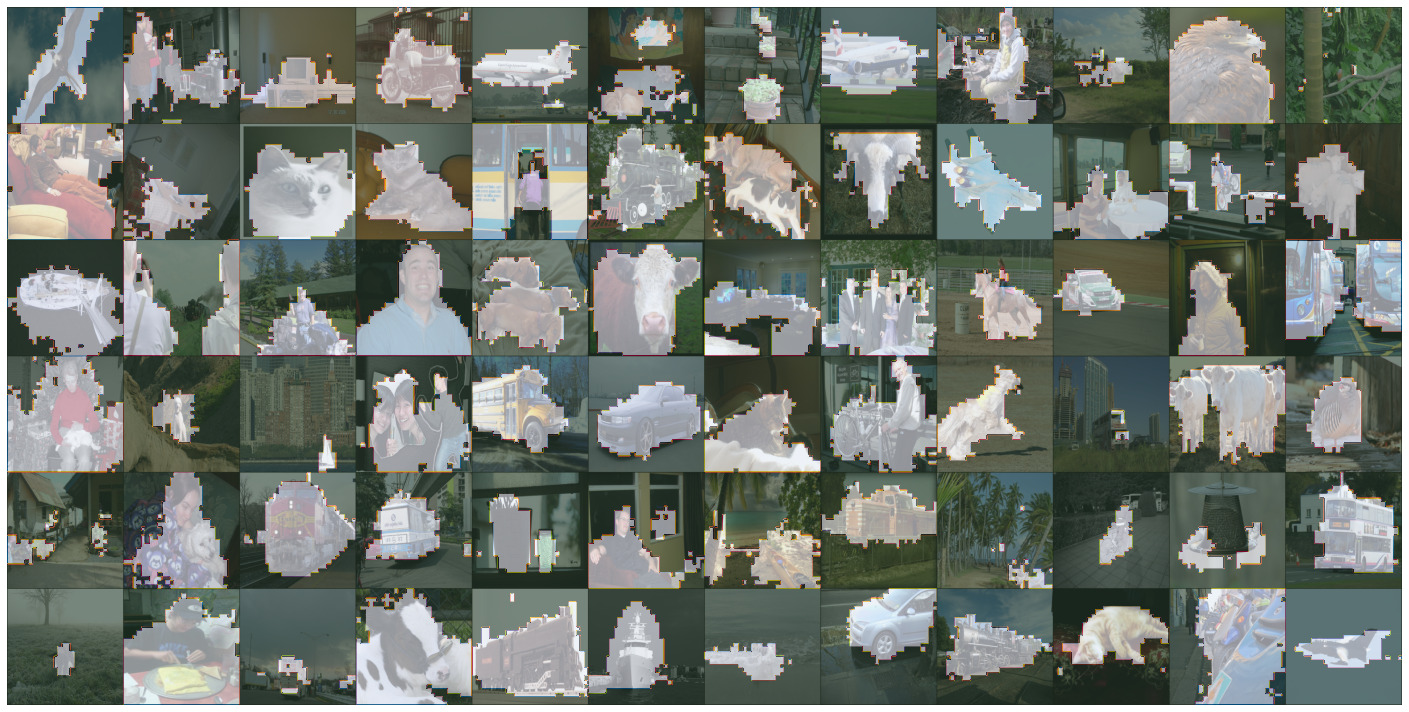}
    \caption{\textbf{More cluster masks for PVOC12 \textit{val} obtained by our CBFE method.} Overall, the masks capture the object shape well but at times they include too much background. Also small objects are not detected at times as can be seen in the first picture from the right in the first row, which is a limitation discussed.}
    \label{fig:more_masks}
\end{figure*}
\paragraph{Evaluation details}
Since we evaluate the pre-GAP \texttt{layer4} features or the spatial tokens, their output resolution does not match the mask resolution. 
To fix that, we bilinearly interpolate before applying the linear or FCN head; or directly interpolate the clustering results by nearest neighbor upsampling.
For a fair comparison between ResNets and ViTs, we use dilated convolution in the last bottleneck layer of the ResNet such that the spatial resolution of both network architectures match (28x28 for 448x448 input images).
All overclustering results were computed using downsampled 100x100 masks to speed up the Hungarian matching as we found that the results do not differ from using full resolution masks.

We fine-tune the linear head for 25 epochs with a learning rate drop after 20 epochs and a batch size of 120.
For most checkpoints we found a learning rate of 0.01 to work well except for the baselines of MaskContrast \cite{vangansbeke2020unsupervised} and MoCo-v2 \cite{he2019momentum} where we use a learning rate of 0.1.
All heads were trained on downsampled 100x100 masks to increase training speed.
For evaluation, we stick to 448x448 masks as it does not require Hungarian matching and is thus fast. 
\par
The FCN head is fine-tuned for 30 epochs equaling the 20k iterations used in \cite{wang2021dense}. 
Again we use a learning rate of 0.01 with a drop to 0.001 after 15 epochs and a batch size of 64.
The design of our fully convolutional head follows \cite{wang2021dense}: We use two convolutional layers with ReLU non-linearites.
Backbone features are fed into the second layer through a skip connection.
The resulting feature maps $\phi$ are then transformed to the desired output classes by a 1x1 convolution.
During training we apply 2D-dropout on $\phi$.

\subsection{Additional Experiments}
\label{sec:ap-additional-experiments}
\paragraph{Fine-Tuning a larger backbone}
To push the boundaries of state-of-the-art even further, we fine-tune a ViT-Base with patch size 8 (ViT-B/8) for 100 epochs with \ours, starting from a DINO initialization.
The results are reported in Tab. 6 in the paper and Tab. \ref{tab:large-back-sota}.
Training a larger backbone boosts transfer learning performance by up to 6\% on PVOC as shown in Tab. \ref{tab:lb-transfer-learning}. 
The gains on COCO-Thing and COCO-Stuff are around $1\%$ and $2\%$ respectively. 
\\
For fully unsupervised semantic segmentation, training a larger backbone even shows more relative gain than training a ViT-Small.
This is apparent from the 41.9\% relative gain for a ViT-B/8 in comparison to the 37.1\% relative gain for a ViT-S/16 over their respective DINO initialization, as can be deduced from Tab. \ref{tab:lb-fullyunsup}.
Overall, we are able to improve state-of-the-art by additional $5.5\%$ just by taking a larger model.
For CD, we found K=110, an edge threshold of $7\%$ and a Markov time of 0.4 to perform best for ViT-B/8. \\
A larger model also improves foreground extraction using our CBFE method by more than $4\%$ as shown in Tab. \ref{tab:lb-fg-extraction}. 

\begin{table}[]
\begin{subtable}[t]{\linewidth}
\centering
    \begin{tabular}{llll}
    \Xhline{2\arrayrulewidth}
    & \multicolumn{3}{c}{K=500} \\
    \cmidrule{2-4}
    arch &  PVOC12 & COCO-Thing & COCO-Stuff \\ \hline
    ViT-S/16 & 53.5 & 55.9 & 43.6  \\ 
    ViT-B/8 & \textbf{59.7} & \textbf{56.8} & \textbf{45.9} \\ 
    \Xhline{2\arrayrulewidth}
    \end{tabular}
\caption{Overclustering results on PVOC, COCO-Thing and COCO-Stuff. The results are comparable to Tab. 3 in the paper.} 
\label{tab:lb-transfer-learning}
\end{subtable}%
\vspace{1em}
\\
\begin{subtable}[t]{\linewidth}
\centering
\begin{tabular}{lll} 
    \Xhline{2\arrayrulewidth}
     & ViT-S/16 & ViT-B/8\\ \hline
    DINO   & 4.6 & \textbf{5.3}\\ 
    + \ours & 18.9 & \textbf{21.2} \\ 
    + CBFE & 36.2 & \textbf{43.3} \\  
    + CD & 41.7 & \textbf{47.2} \\ \hline
    \Xhline{2\arrayrulewidth}
\end{tabular}
\caption{Fully unsupervised semantic segmentation results on PVOC.}
\label{tab:lb-fullyunsup}
\end{subtable}
\vspace{1em}
\\
\begin{subtable}[t]{\linewidth}
\centering
\begin{tabular}{lll}
\Xhline{2\arrayrulewidth}
Method      & arch &  Jacc. (\%)  \\ \hline
\ours\ IN CBFE & ViT-S/16 & 58.6 \\ 
\ours\ CC CBFE & ViT-S/16 & 59.6 \\
\ours\ CC CBFE & ViT-B/8 & \textbf{63.5} \\ \Xhline{2\arrayrulewidth}
\end{tabular}
\caption{Foreground extraction results on PVOC. The results are comparable to Table 7 in the paper.}
\label{tab:lb-fg-extraction}
\end{subtable}
\vspace{1em}
\caption{\textbf{Comparison of ViT-S/16 and ViT-B/8 performances.} We further improve state-of-the-art on all experiments by training a larger model with our loss and running CBFE and CD.}
\label{tab:large-back-sota}
\end{table}

\begin{table}[h]
\footnotesize
    \centering
    \begin{tabular}{lc cc cc}
    \toprule
         & & \multicolumn{2}{c}{\textbf{At init.}} &  \multicolumn{2}{c}{\textbf{After Leopart}}  \\
         Init & Arch & LC & K=500 & LC & K=500 \\
         \hline 
         Superv. & ViT-S/16 & 68.1 & 55.1 & 72.5 & 61.6 \\
         MoCo-v3~\cite{chen2021mocov3} & ViT-S/16 & 13.4 & 5.8 & 42.0 & 31.2 \\
         MAE~\cite{mae} & ViT-B/16 & 47.5 & 10.0 & 68.9 & 38.4 \\
         \bottomrule
    \end{tabular}
    \caption{\textbf{Transfer learning results starting from various initializations}. \ours\ consistently improves upon the initialization (init.) and thus shows the generality of our method. Comparable to Tab. 3 in the paper.} 
    \label{tab:other-backbone}
\end{table}
\paragraph{\ours\ with different initializations}
To show the generality and robustness of our approach, we fine-tune with \ours\ starting from a Moco-v3, MAE and supervised initialization.
The results are shown in Table \ref{tab:other-backbone}.
\ours\ is good at fine-tuning even more recent SSL methods and larger pretrained backbones like MAE (where our method adds $+28\%$ in K=500 performance).
Our method is even able to boost the performance of a ViT pretrained with supervision showing the wide applicability of our dense loss.

\paragraph{DenseCL with DINO init.}
For further comparison to our closest competitor in transfer learning, DenseCL~\cite{wang2021dense}, we trained a ViT with DINO initalization using their loss and following the setting of Tab.~3 for PVOC12. 
We find a performance of {54\% and 17.1\%} for LC and K=500 evaluation respectively, \ie fine-tuning with \ours\ still outperforms by {$>\!\!15\%$ for LC and $>\!\!40\%$ for K=500}.
These results indicate that DenseCL (perhaps due to its global-pooled loss term) does not seem apt for fine-tuning as it barely improves upon the DINO initialisation ($+3.4\%$ for LC and $-0.3\%$ for K=500).

\paragraph{Queue usage ablation.}
\begin{table}[h]
\centering
    \begin{tabular}{lllll}
    \Xhline{2\arrayrulewidth}
    &&  \multicolumn{3}{c}{Num. clusters} \\
    \cmidrule{3-5}
    queue & LC & 100 & 300 & 500 \\ \hline
    \xmark  & 67.2 & 35.0 & 45.7 & 48.1  \\ 
    \checkmark & \textbf{67.8} & \textbf{38.2} & \textbf{47.2} & \textbf{50.7} \\ 
    \Xhline{2\arrayrulewidth}
    \end{tabular}
\caption{\textbf{Queue Ablation}. A clustering queue improves performance.}
\label{tab:queue-ablation}
\end{table}
We show that the usage of a queue improves our results as shown in Table \ref{tab:queue-ablation}, comparable to the experiments of Table 1 in the main paper.
This means that enough diversity for equi-partitioned clustering can be achieved with this simple mechanism.

\begin{table}[]
\centering
\begin{tabular}{l|ll} \Xhline{2\arrayrulewidth}
\multirow{2}{*}{Method} & \multicolumn{2}{c}{ADE20k-Street} \\ 
& K=500 & K=1000\\ \hline
Random ViT & 1.5 & 2.0 \\
Sup. ViT  & 5.4 & 7.2  \\
DINO \cite{caron2021emerging} & 5.7 & 7.0 \\
\ours\ IN & \underline{6.9} & \underline{9.3}\\
\ours\ CC & \textbf{7.6} & \textbf{10.0} \\ \Xhline{2\arrayrulewidth}
\end{tabular}
\caption{\textbf{ADE20k overclustering results.} Evaluated on 111 parts classes taken from ADE20k street scenes.}
\label{tab:ade20kparts}
\end{table}
\paragraph{Predicting ADE20k parts}
To quantitatively support our claim that we learn object parts, we run experiments on ADE20K \cite{zhou2017scene} street scenes that feature annotations for 111 different part classes on 1983 images.
We pretrain on COCO and report overclustering results given ground-truth parts annotations. 
As shown in Table \ref{tab:ade20kparts},  \ours\ improves DINO's parts mIoU by $1.9\%$ and $3\%$ with a clustering granularity of $K=500$ and $K=1000$ respectively.
This shows that our method increases object part correspondence.
Also, while the supervised ViT outperformed DINO in transfer learning it is not superior when it comes to discovering object parts.

\begin{figure*}[t]
    \centering
    \includegraphics[width=\textwidth]{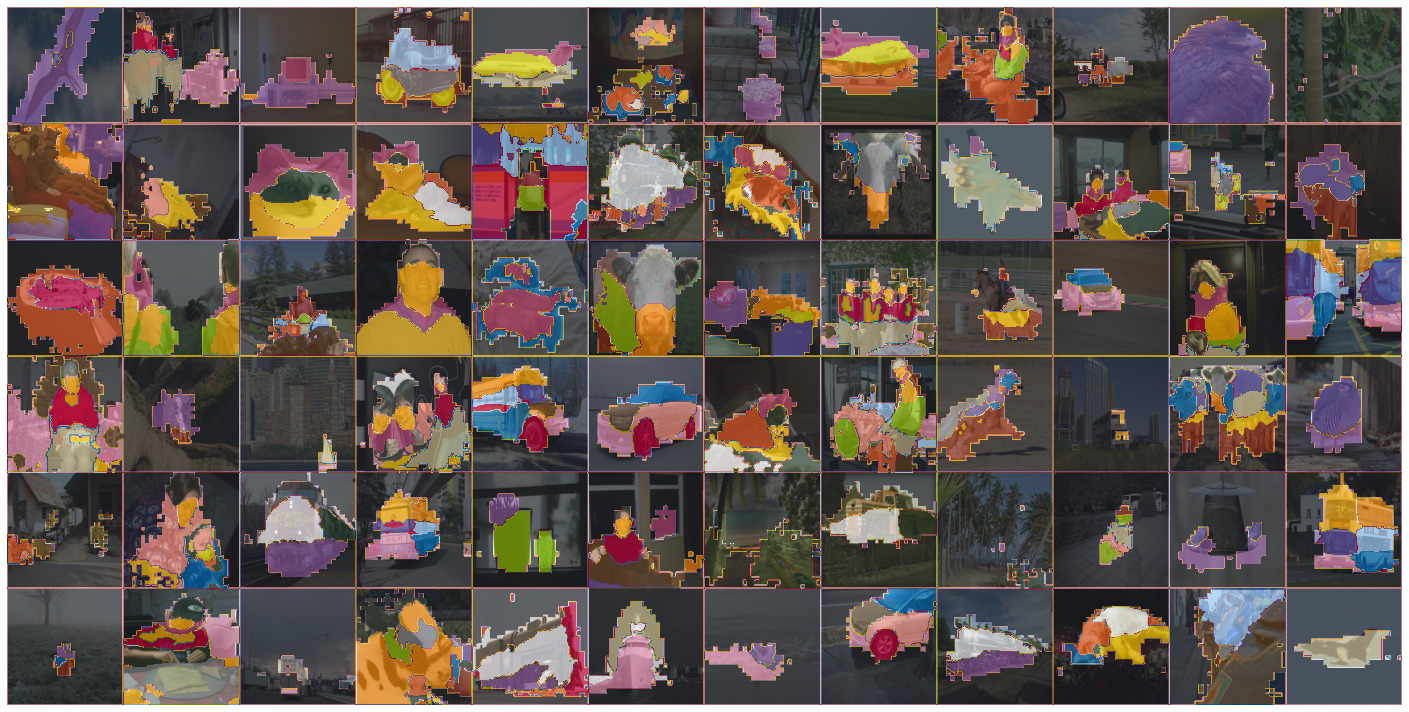}
    \caption{\textbf{K=100 overclustering visualization without merging clusters to objects.} Note that the cluster colors are not unique as we have 100 different clusters: Same cluster means same color but not the other way around. Interestingly, \ours\ learns a different segmentation granularity depending on the object category. For instance, cars and humans are segmented into various parts, but birds are usually kept whole.}
    \label{fig:k100-clusters}
\end{figure*}
\begin{figure*}[h]
    \centering
    \includegraphics[width=\textwidth]{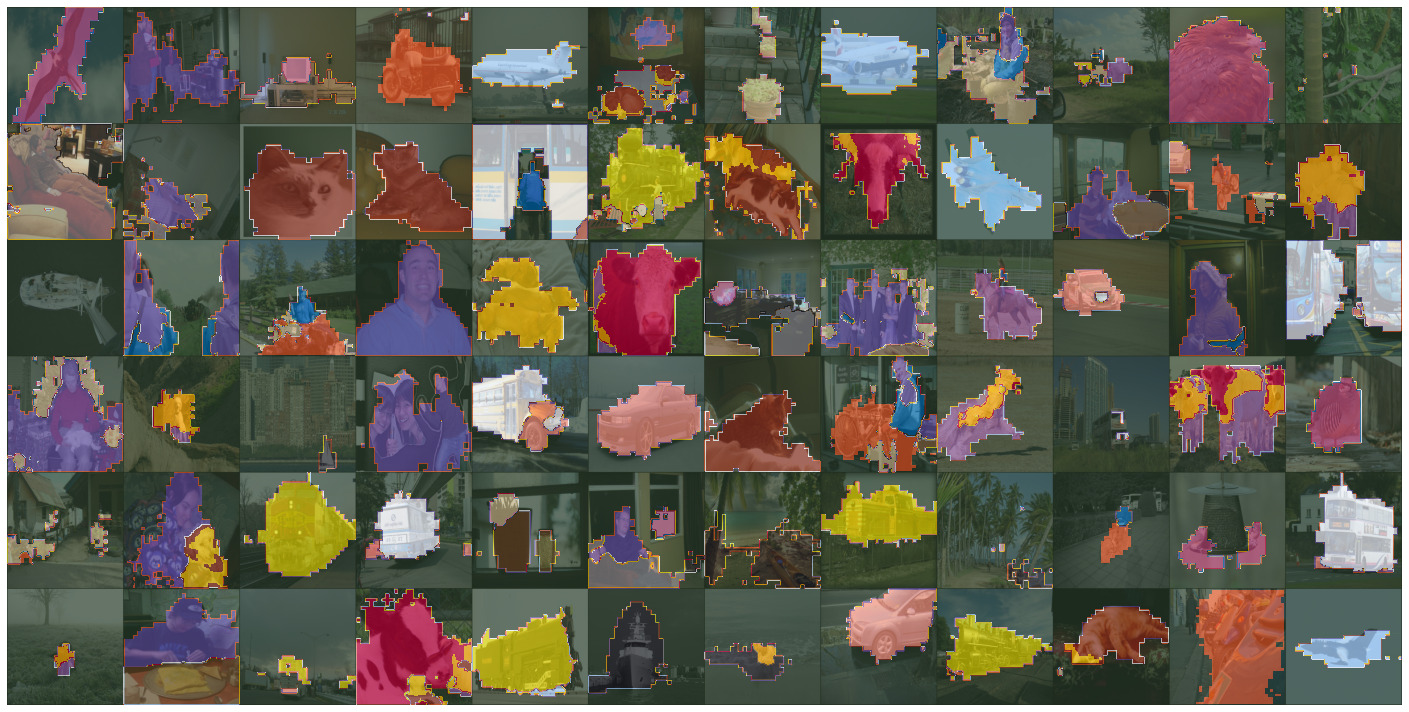}
    \caption{\textbf{More fully unsupervised segmentation results obtained through our community detection method.} Our method, manages to merge the object parts clusters from Figure \ref{fig:k100-clusters} to objects in most of the cases. However, as our method does a hard cluster to community assignment, each cluster can only be used for one object. This limitation can be seen for the car wheel class in the 4th row and 5th and 6th pictures from the right. The bus' wheel is mistakenly assigned to the car category. Also, objects that share many parts such as bicycles and motorcycles are mistakenly merged to one category.}
    \label{fig:more_cd}
\end{figure*}
\begin{figure*}[h]
    \centering
    \includegraphics[width=\textwidth]{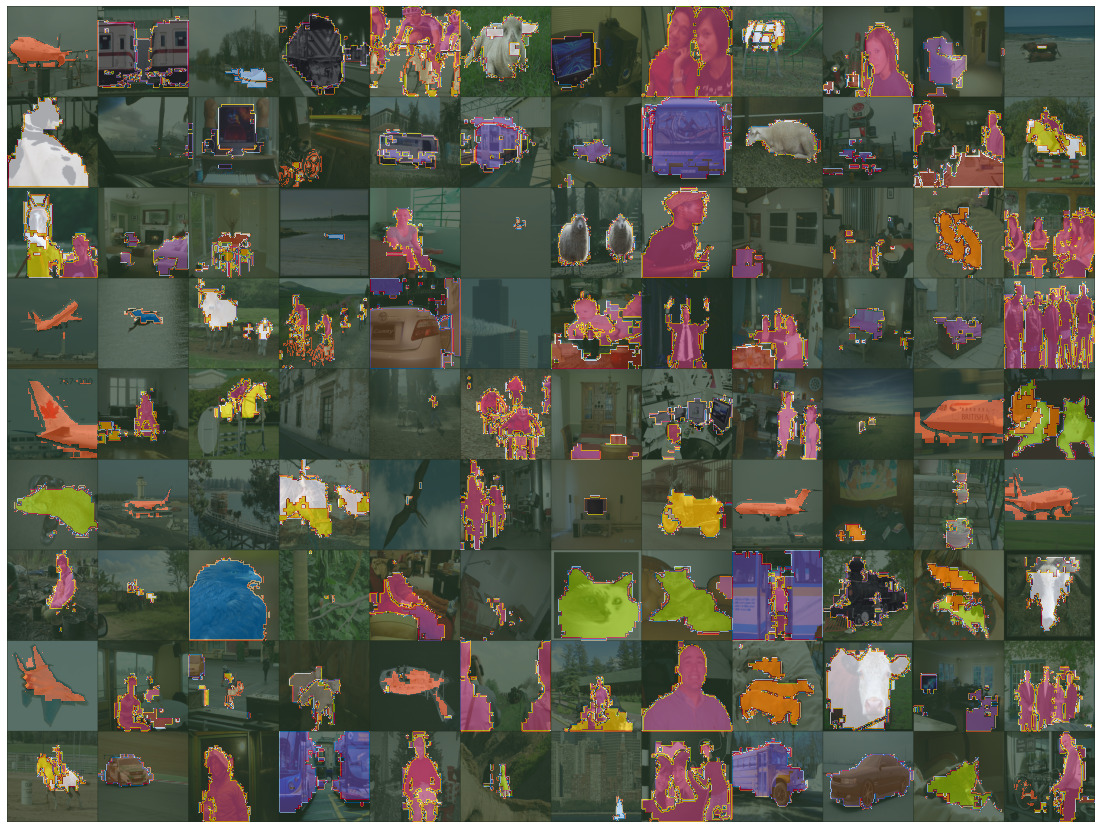}
    \caption{\textbf{Overclustering results by merging 500 clusters using ground-truth labels.} The resulting segmentation maps are more crisp than their fully unsupervised counterparts in Figure \ref{fig:more_cd}. Further, similar object categories are not merged together. However at times, the object is not fully segmented but just parts of it. This is likely due to the fact that some clusters segmenting an object also have a significant background overlap and thus our precision-based cluster matching matches them to the background class.}
    \label{fig:500-merged}
\end{figure*}
\subsection{Additional visualizations}
We provide further cluster masks in Figure \ref{fig:more_masks} and segmentation map visualizations on PVOC12.
Next to community detection results shown in Figure \ref{fig:more_cd}, we also show unmerged foreground clustering results with K=100 in Figure \ref{fig:k100-clusters} to give the reader an impression of the segmentation granularities of each object.
In Figure \ref{fig:500-merged}, we also show segmentation maps obtained from classic overclustering results by grouping clusters to objects using label information.

\subsection{Datasets Details}
\subsubsection{PASCAL}
For fine-tuning linear heads as well as the FCN head, we use the \textit{trainaug} split featuring 10582 images and their annotations.
We evaluate on PVOC12 \textit{val} that has 1449 images.
During evaluation we ignore unlabelled objects as well and the boundary class following \cite{vangansbeke2020unsupervised}.
For hyperparameter tuning of our fully unsupervised segmentation method, we use the PVOC12 \textit{train} split with 1464 images.

\subsection{COCO}
\begin{figure*}[h]
     \begin{subfigure}[b]{0.495\textwidth}
         \centering
         \includegraphics[width=\textwidth]{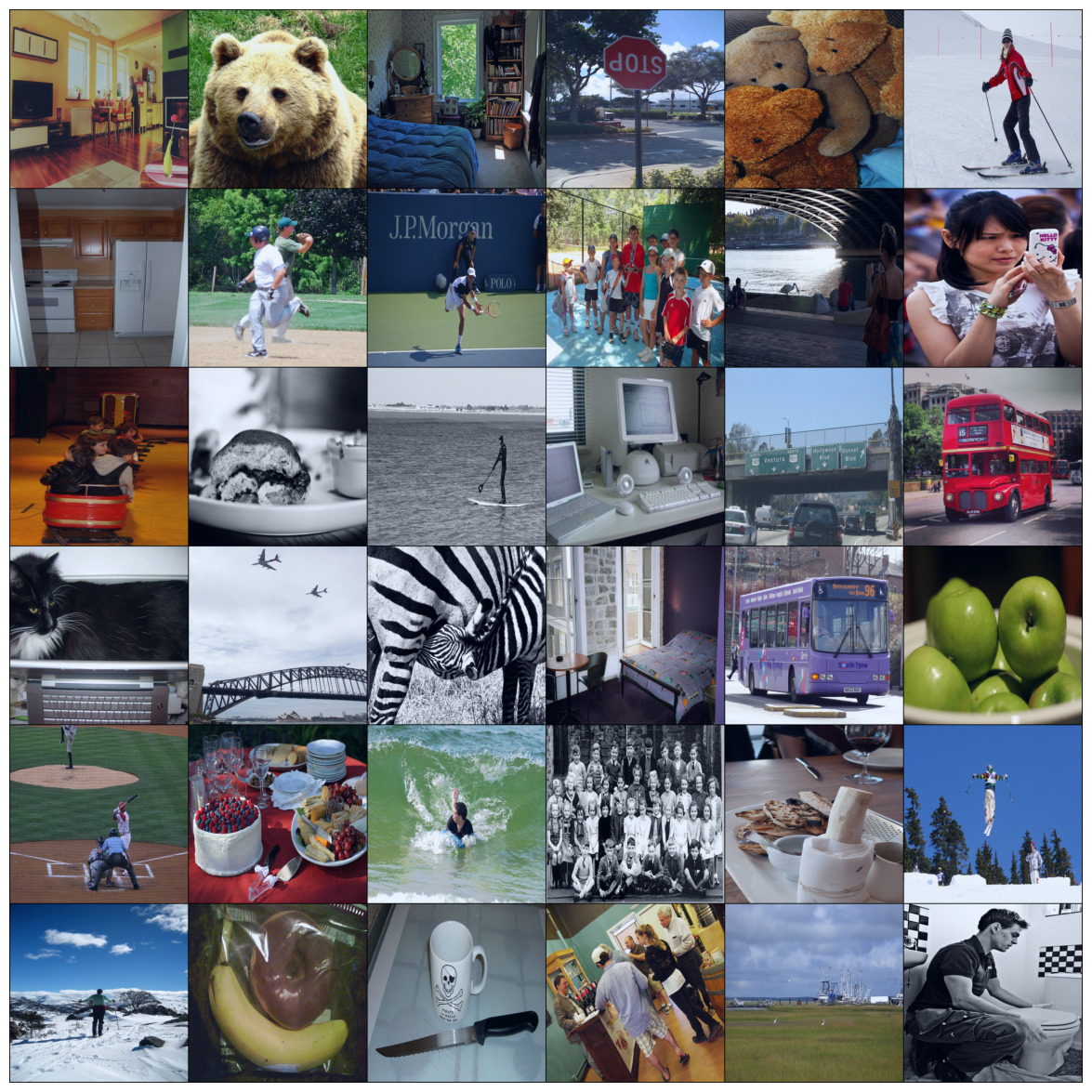}
         \caption{COCO stuff images}
     \end{subfigure}%
     \hfill
     \begin{subfigure}[b]{0.495\textwidth}
         \centering
         \includegraphics[width=\textwidth]{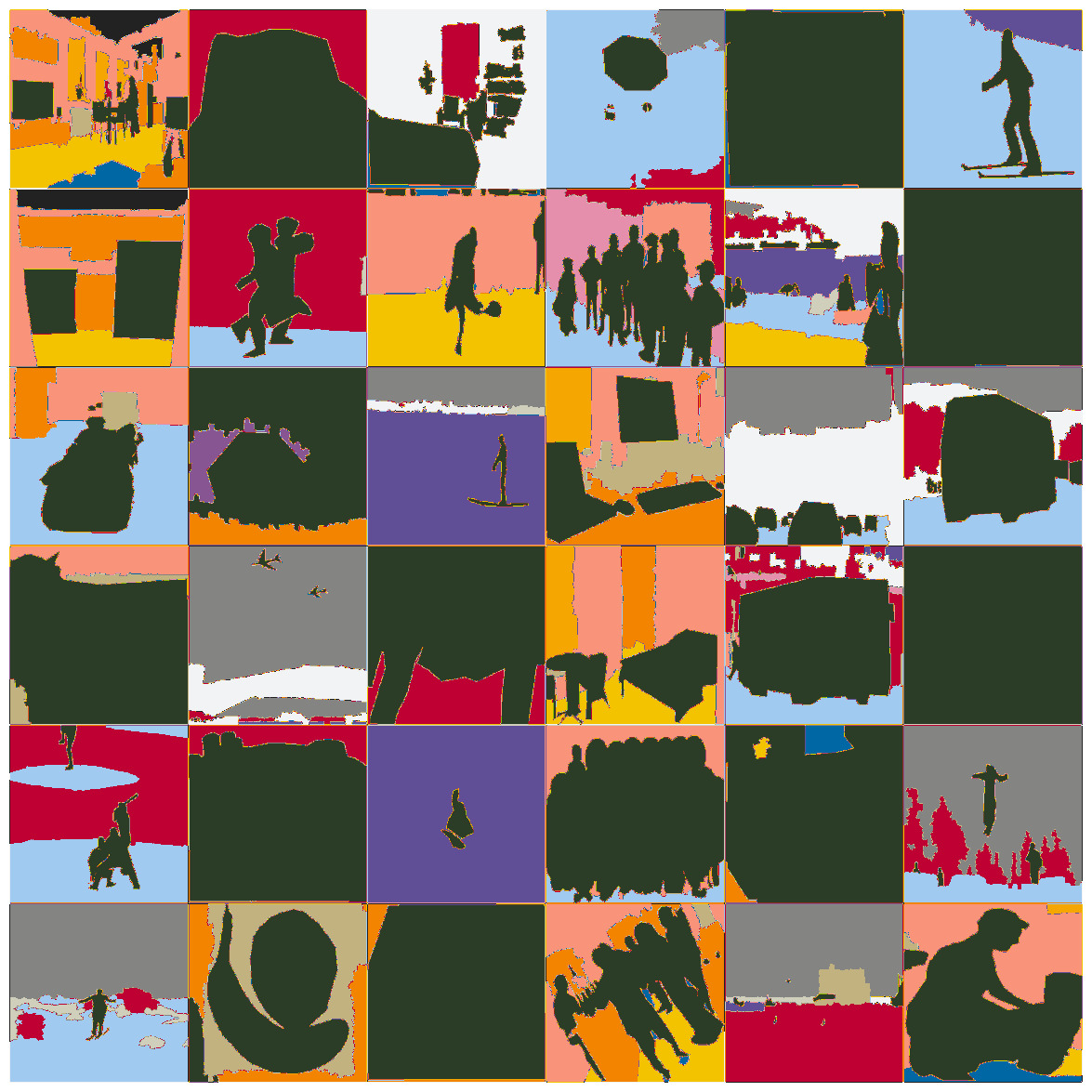}
         \caption{COCO stuff masks}
     \end{subfigure}%
     \hfill
    \caption{\textbf{COCO-stuff images and masks data visualized as used for head fine-tuning and evaluation.} Some pictures are completely dark green and thus ignored as the stuff class appearing is ``other''.}
    \label{fig:stuff-data}
\end{figure*}
\begin{figure*}[h]
     \begin{subfigure}[b]{0.495\textwidth}
         \centering
         \includegraphics[width=\textwidth]{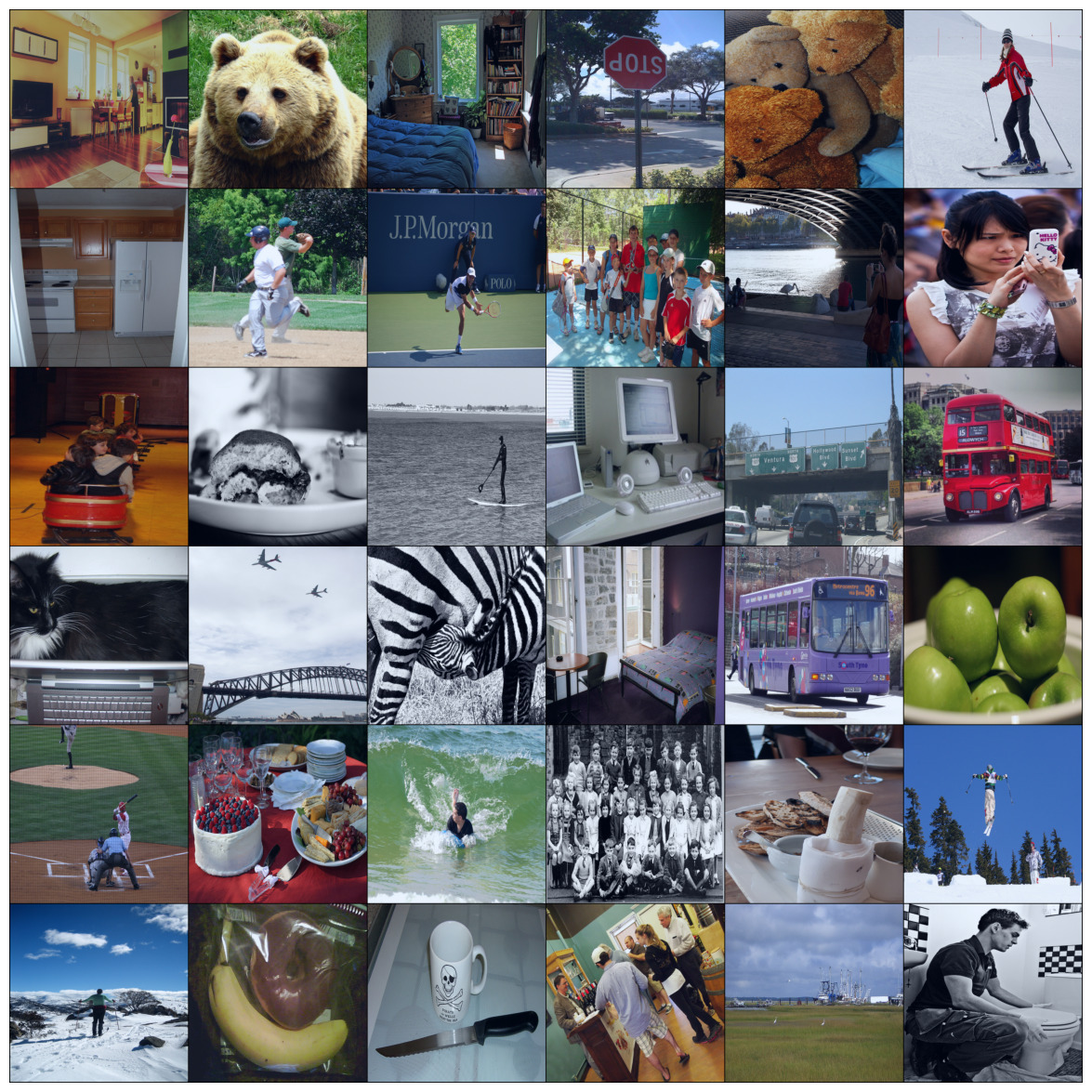}
         \caption{COCO Things Images}
     \end{subfigure}%
     \hfill
     \begin{subfigure}[b]{0.495\textwidth}
         \centering
         \includegraphics[width=\textwidth]{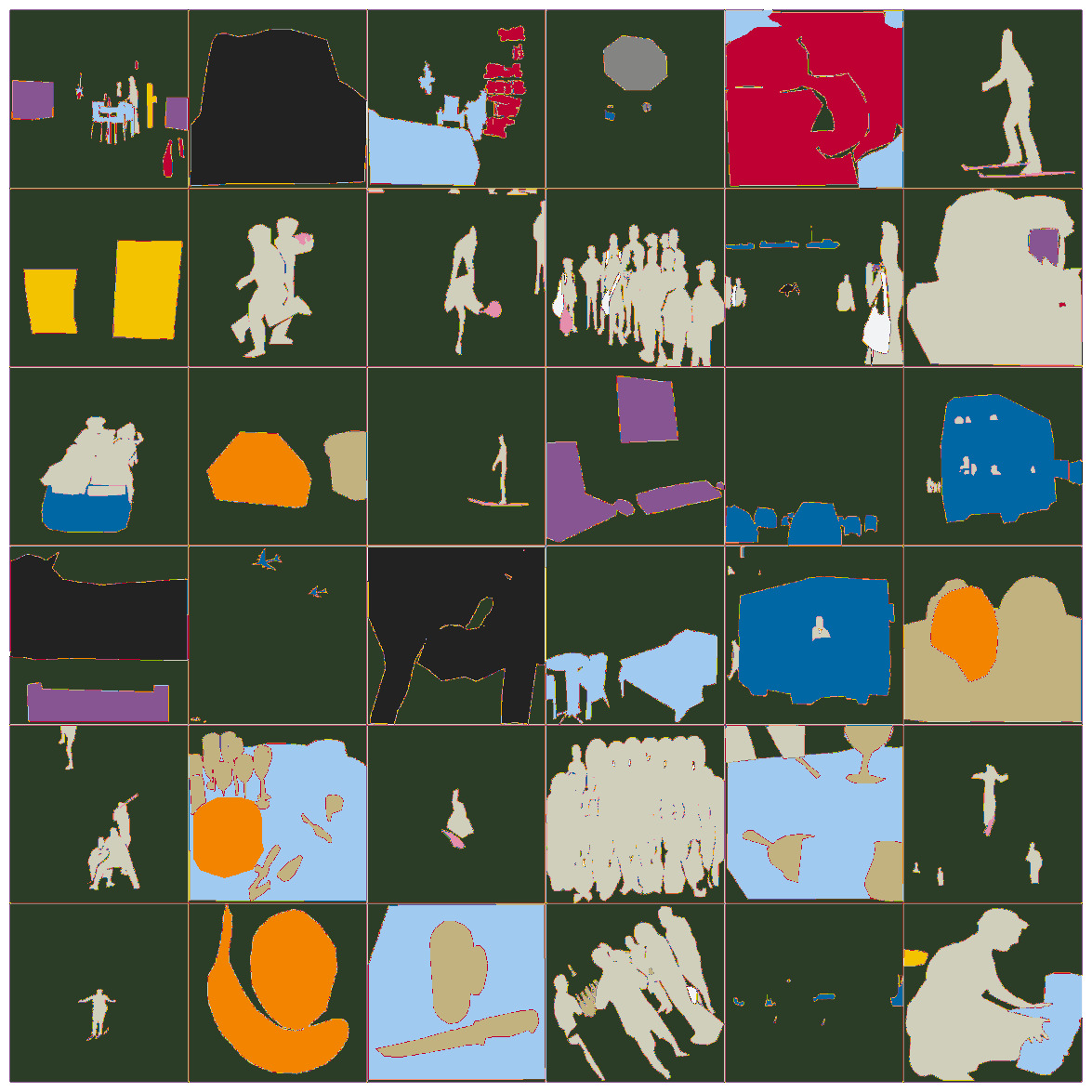}
         \caption{COCO Things masks}
     \end{subfigure}%
     \hfill
    \caption{\textbf{COCO-thing images and masks data visualized as used for head fine-tuning and evaluation.}}
    \label{fig:thing-data}
\end{figure*}
We use the COCO benchmark in two ways to further isolate different object definitions.
For instance, COCO-thing has one class for vehicles whereas PVOC distinguishes between boats, busses and cars.
Also, things have a fundamentally different object definition as stuff.
First, we extract stuff annotations i.e. object w/o a clear boundary, often in the background.
For that, we use the COCO-Stuff annotations \cite{caesar2018coco}.
We further merge the 91 fine labels to 15 coarse labels, as in \cite{ji2018invariant}.
We also assign the coarse label ``other'' to non-stuff object as the label does not carry any semantic meaning.
The resulting labels are:
\begin{verbatim}
    ['water', 'structural', 'ceiling', 
    'sky', 'building', 'furniture-stuff', 
    'solid', 'wall', 'raw-material', 
    'plant', 'textile', 'floor', 
    'food-stuff', 'ground', 'window']
\end{verbatim}
Non-Stuff objects are ignored during training and evaluation. Figure \cref{fig:stuff-data} shows some exemplary coco-stuff images and their corresponding masks.
\par
Second, we extract foreground annotations by using the panoptic labels provided by \cite{kirillov2019panoptic}.
We merge the instance-level annotations to an object category with a script the authors provided.
Further, we merge the 80 fine categories to coarse categories obtaining 12 unique object classes:
\begin{verbatim}
    ['electronic', 'kitchen', 'appliance', 
    'sports', 'vehicle', 'animal', 
    'food', 'furniture', 'person', 
    'accessory', 'indoor', 'outdoor']
\end{verbatim}
The background class is ignored during training and evaluation. Figure \cref{fig:thing-data} shows some exemplary coco-thing images and their corresponding masks.
\par
We fine-tune the linear and FCN head on a subset of 10\% of the data i.e. 11829 images.
We evaluate on the full 5000 validation images.

\begin{figure*}[h]
     \begin{subfigure}[b]{0.495\textwidth}
         \centering
         \includegraphics[width=\textwidth]{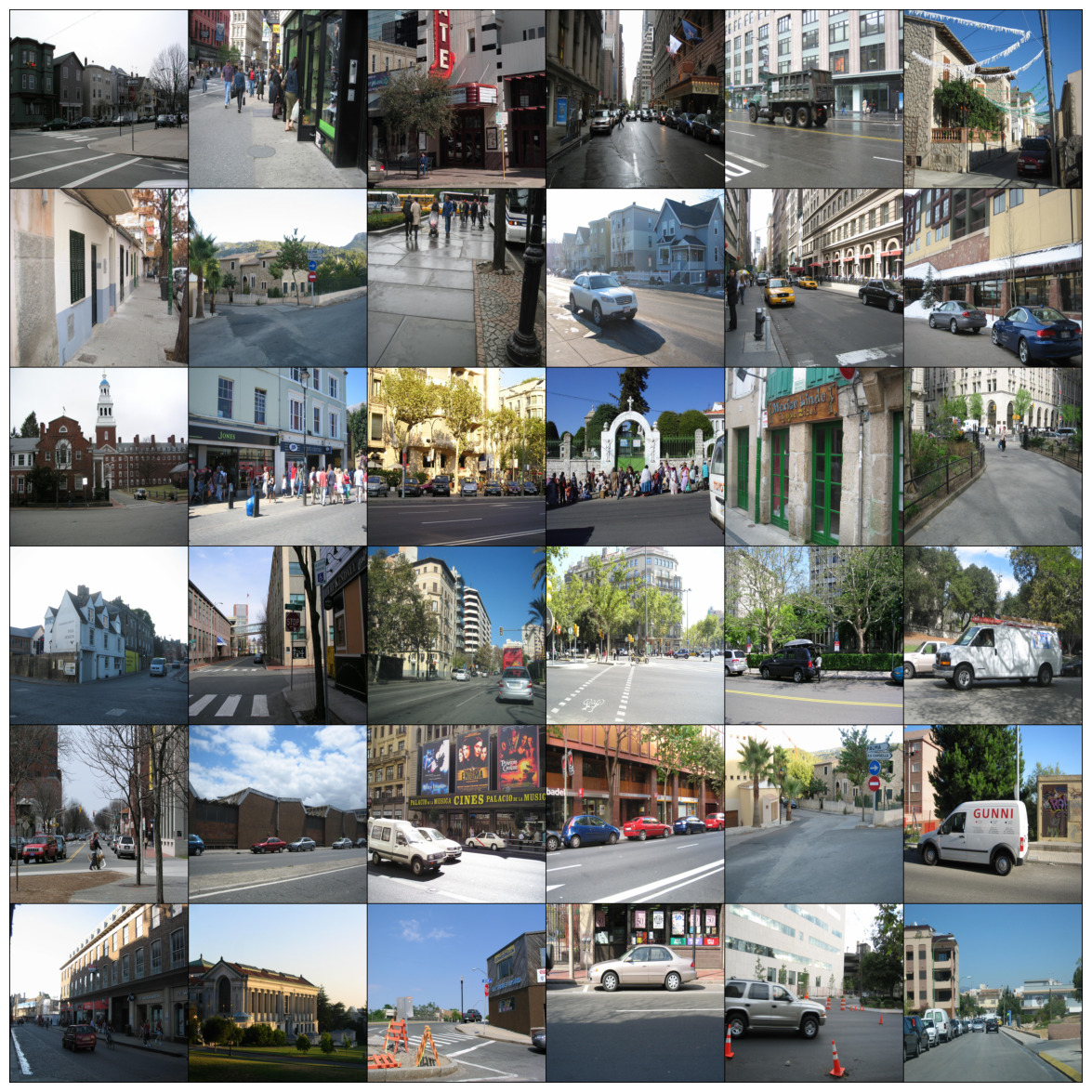}
         \caption{ADE20k street scene images}
     \end{subfigure}%
     \hfill
     \begin{subfigure}[b]{0.495\textwidth}
         \centering
         \includegraphics[width=\textwidth]{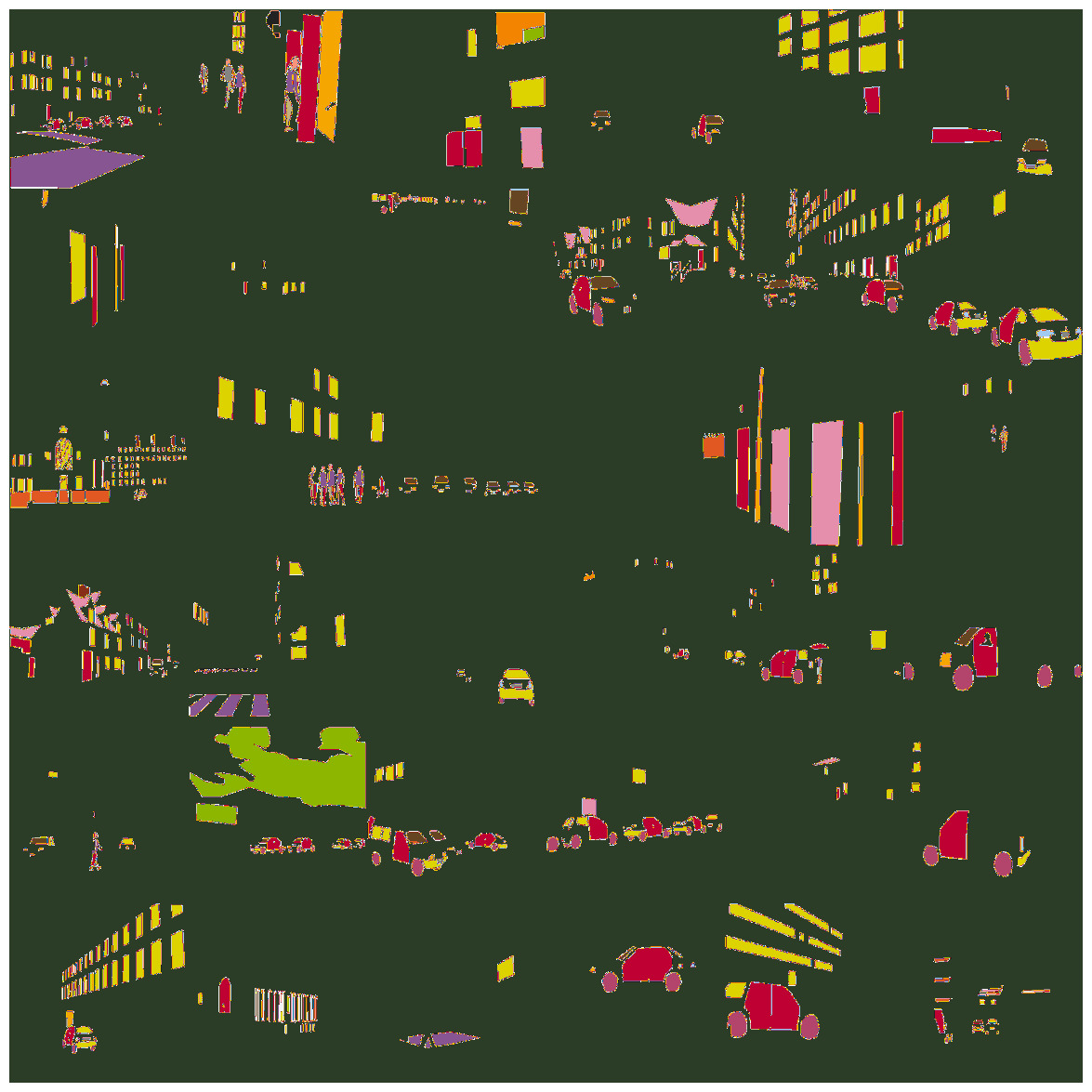}
         \caption{ADE20k street scene parts masks}
     \end{subfigure}%
     \hfill
    \caption{\textbf{ADE20k street scene images and masks data visualized as used for overclustering results reported in Table \ref{tab:ade20kparts}.}}
    \label{fig:20k-data}
\end{figure*}
\subsubsection{ADE20k}
Overall, ADE20k features 3687 different objects that can act as parts. 
We constrain our evaluation to street scenes that contain parts annotations.
This reduces our data to 1983 images and to the following 111 object parts:
\begin{verbatim}
    ['arcades', 'arch', 'arm', 
    'back', 'balcony', 'balustrade', 
    'bars', 'base', 'basket', 
    'bell', 'bicycle path', 'blind', 
    'blinds', 'branch', 'bumper', 
    'chimney', 'cloud', 'clouds', 
    'column', 'columns', 'cornice', 
    'crosswalk', 'dome', 'door', 
    'door frame', 'doorbell', 'dormer', 
    'double door', 'drain pipe', 'eaves', 
    'entrance', 'entrance parking', 
    'exhaust pipe', 'face', 'fence', 
    'fender', 'fire bell', 'fire escape', 
    'garage door', 'garage doors', 
    'gas cap', 'gate', 'grille', 
    'ground', 'gutter', 'handle', 'head', 
    'headboard', 'headlight', 'hip tiles', 
    'hood', 'house number', 'housing', 
    'housing lamp', 'lamp', 
    'lamp housing', 'lattice', 'left arm', 
    'left foot', 'left hand', 'left leg', 
    'license plate', 'logo', 
    'metal shutter', 'metal shutters', 
    'mirror', 'pipe', 'pipe drain', 
    'pole', 'porch', 'post', 'railing', 
    'rain pipe', 'revolving door', 
    'right arm', 'right foot', 
    'right hand', 'right leg', 'rim', 
    'road', 'roof', 'roof rack', 
    'rose window', 'saddle', 
    'shop window', 'shutter', 'shutters', 
    'sidewalk', 'sign', 'skylight', 
    'staircase', 'steering wheel', 
    'step', 'steps', 'taillight', 
    'terrace', 'torso', 'tower', 'tree', 
    'trunk', 'vent', 'wall', 'wheel', 
    'window', 'window scarf', 'windows', 
    'windshield', 'wiper', 'car', 
    'buildings', 'building']
\end{verbatim}
Figure \ref{fig:20k-data} shows some exemplary street scene images and their corresponding parts masks.
During evaluation of our feature space clustering we ignore non-part pixels shown in dark green.

\end{document}